\documentclass{article} 

\usepackage{amsmath,amsfonts,bm}

\def\figref#1{figure~\ref{#1}}

\def\secref#1{section~\ref{#1}}

\def\eqref#1{equation~\ref{#1}}

\def\algref#1{algorithm~\ref{#1}}

\def\1{\bm{1}}

\DeclareMathAlphabet{\mathsfit}{\encodingdefault}{\sfdefault}{m}{sl}
\SetMathAlphabet{\mathsfit}{bold}{\encodingdefault}{\sfdefault}{bx}{n}

\newcommand{\E}{\mathbb{E}}

\usepackage{hyperref}
\usepackage{url}
\usepackage{enumitem,kantlipsum}

\usepackage[margin=1in]{geometry}

\usepackage{amsthm,amsfonts,amsmath,amssymb,epsfig,color,float,graphicx,verbatim}
\usepackage[ruled,vlined]{algorithm2e}
\usepackage[authoryear]{natbib}
\usepackage{wrapfig}
\usepackage{authblk}

\usepackage{hyperref}
\hypersetup{
	colorlinks   = true, 
	urlcolor     = blue, 
	linkcolor    = blue, 
	citecolor   = black 
}
\usepackage{bbm}
\usepackage{newtxtext}
\usepackage[varbb]{newtxmath}
\usepackage{caption}
\usepackage{subcaption}
\usepackage{enumitem}
\usepackage{multicol}

\newtheorem{theorem}{Theorem}[section]

\newtheorem{corollary}[theorem]{Corollary}

\newcommand{\reals}{\mathbb{R}}

\newcommand{\var}{\text{Var}}

\newcommand{\ifrac}[2]{#1\smash{\big/}#2}

\newcommand{\be}{\mathbf{e}}
\newcommand{\bx}{\mathbf{x}}

\newcommand{\bg}{\mathbf{g}}

\newcommand{\bd}{\mathbf{d}}
\newcommand{\bh}{\mathbf{h}}

\newcommand{\bxi}{\boldsymbol{\xi}}
\newcommand{\bzeta}{\boldsymbol{\zeta}}

\newcommand{\btheta}{\boldsymbol{\theta}}

\newcommand{\Ncal}{\mathcal{N}}

\newcommand{\norm}[1]{\|#1\|}
\newcommand{\inner}[1]{\langle#1\rangle}

\renewcommand{\secref}[1]{Section~\ref{#1}}

\renewcommand{\eqref}[1]{Eq.~(\ref{#1})}

\newcommand{\corollaryref}[1]{Corollary~\ref{#1}}
\newcommand{\thmref}[1]{Thm.~\ref{#1}}

\newcommand{\appref}[1]{Appendix~\ref{#1}}

\makeatletter
\newcommand{\printfnsymbol}[1]{%
  \textsuperscript{\@fnsymbol{#1}}%
}
\makeatother

\title{Locally Optimal Descent for Dynamic Stepsize Scheduling}

\author[1, 2]{Gilad Yehudai}
\author[1, 3]{Alon Cohen}
\author[1, 4]{Amit Daniely}
\author[1]{Yoel Drori}
\author[1, 5]{Tomer Koren}
\author[1]{Mariano Schain}

\affil[1]{\small Google Research Tel Aviv}
\affil[2]{\small Weizmann Institute of Science}
\affil[3]{\small School of Electrical Engineering, Tel Aviv University}

\affil[4]{\small Hebrew University of Jerusalem}
\affil[5]{\small Blavatnik School of Computer Science, Tel Aviv University}

\date{}
\begin{document}
\maketitle

\begin{abstract}
We introduce a novel dynamic learning-rate scheduling scheme grounded in theory with the goal of simplifying the manual and time-consuming tuning of schedules in practice.  Our approach is based on estimating the locally-optimal stepsize, guaranteeing maximal descent in the direction of the stochastic gradient of the current step.  We first establish theoretical convergence bounds for our method within the context of smooth non-convex stochastic optimization, matching state-of-the-art bounds while only assuming knowledge of the smoothness parameter.  We then present a practical implementation of our algorithm and conduct systematic experiments across diverse datasets and optimization algorithms, comparing our scheme with existing state-of-the-art learning-rate schedulers. Our findings indicate that our method needs minimal tuning when compared to existing approaches, removing the need for auxiliary manual schedules and warm-up phases and achieving comparable performance with drastically reduced parameter tuning.
\end{abstract}

\section{Introduction}
Stochastic gradient-based optimization methods such as SGD and Adam~\citep{kingma2014adam} are the main workhorse behind modern machine learning.
Such methods sequentially apply stochastic gradient steps to update the trained model and their performance crucially depends on the choice of a learning rate sequence, or schedule, used throughout this process to determine the magnitude of the sequential updates. 
All in all, effectively tuning the learning rate schedule is widely considered a tedious task requiring extensive, sometimes prohibitive, hyper-parameter search, resulting in a significant excess of engineering time and compute resources usage in ML training. 

A prominent approach to address this issue gave rise to a plethora of \emph{adaptive optimization methods} (most notably \citealp{duchi2011adaptive} and \citealp{kingma2014adam}), where the learning rate parameter is automatically tuned during the optimization process based on previously received stochastic gradients. 
In some important applications these methods provide superior convergence performance, while their theoretical guarantees match the state-of-the-art in the stochastic convex and (smooth) non-convex optimization settings~\citep{li2019convergence,ward2020adagrad,pmlr-v202-attia23a}.
However, despite the adaptivity incorporated into these methods, 
auxiliary learning rate schedules are often still required to actually attain their optimal performance~(e.g.,~\citealp{loshchilov2016sgdr}), and the nuisance of laborious and extensive manual tuning still remain relevant for these methods as well.
Furthermore, and perhaps more fundamentally, commonly used schedules appear rather arbitrary and manually tailored to the specific task at hand, and it is desirable to have a more principled, theory-grounded and general-purpose approach to schedule tuning.
%

In this paper, we introduce a novel dynamic learning-rate scheduling scheme, we call {\em GLyDER},\footnote{GLyDER stands for Greedy Local Descent optimizER.} with the goal of automatizing and simplifying the manual and time-consuming tuning of schedules in practice.
GLyDER is principled on an update rule that uses a locally-optimal learning rate, that is, a step-size picked so as to optimize (a bound on) the achievable single-step (greedy) improvement in function value.
This apparatus can be used as an add-on on top of any first-order optimization method that provides (stochastic) step directions for which locally-adapted learning rates are desired.
Resulting algorithms from this scheme do not require external learning-rate schedules and are shown to achieve comparable performance, across several tasks and datasets, to state-of-the-art optimizers that rely on carefully tuned learning rate schedules.

The basic approach behind GLyDER is simple and lies on the classical ``descent lemma'' (e.g.,~\citealp{bauschke10convex,beck2017first}), that quantifies a worst-case bound on the achievable single-step improvement in a given direction, which in turn dictates a choice of a learning rate that optimizes the bound.  
In GLyDER, the idea is to approximate this optimal learning rate given that the direction of progress is the \emph{current stochastic gradient} (possibly mini-batched, or otherwise modified), rather than the true non-stochastic gradient as in classical uses of the descent lemma.
This approximation requires methods for effectively estimating the true gradient's norm and its inner product with the stochastic gradient, based solely on additional samples of stochastic gradients. We give rigorous methods to accomplish these tasks, along theoretical analysis and more practical variants used for our actual implementation of GLyDER which is used in our experiments.


To summarize, our main contributions in this paper are as follows:

\begin{itemize}[leftmargin=4ex]
\item We present a new dynamic scheme for learning-rate scheduling, based on estimating the locally-optimal step-size \emph{in the direction of the current-step stochastic gradient}; the ideas underlying this scheme and its precise derivation are described in \secref{sec:optimal learning rate}.
In contrast to common adaptive stochastic optimization methods, our scheme is able to adapt the instantaneous learning rate to the \emph{current, local} conditions, rather than to \emph{past} observations.

    
\item We prove theoretical convergence bounds for our method in the smooth, non-convex stochastic optimization setting, while only requiring knowledge of the smoothness parameter.  Our bounds match the state-of-the-art performance known for perfectly-tuned SGD in the smooth, non-convex setting~\citep{ghadimi2013stochastic}, and are discussed in \secref{sec:optimal learning rate}.

    
\item We propose two practical implementation variants of of the method and experiment with them extensively across several datasets and optimizers.  Our findings, described in \secref{sec:experiments}, demonstrate that our method achieves performance comparable to state-of-the-art learning rate schedulers.  Notably, our method requires having to tune only a single learning rate parameter, in contrast to other schemes that often require extensive parameter tuning, manually crafted schedules and warm-up phases.

\end{itemize}










\subsection{Related work}

\paragraph{Adaptive stochastic optimization.} 

A closely related and extremely influential line of work, with similar motivation to ours, is that on so-called adaptive optimization methods.  This body of research has originated in the online learning literature and has lead to an abundance of practical and effective optimizers, some of which are used extensively in modern training pipelines~\citep{duchi2011adaptive,kingma2014adam,loshchilov2017decoupled,gupta2018shampoo,shazeer2018adafactor,anil2020scalable,ward2020adagrad}.
Our study complements this family of algorithms in two important aspects: (i) our approach is based on optimizing the instantaneous learning rate to the local conditions in the \emph{current} step, which is very different from the purpose of adaptive optimizers that adapt the learning rate based on \emph{past} gradients; (ii) as already noted above, existing adaptive optimizers often still require auxiliary, manually-tuned learning rate schedules, and our method can be used in conjunction so as to provide them so as to provide them an appropriate schedule.

\paragraph{Stochastic line-search methods.}

Several works \citep{schaul2013no,wu2018understanding,rolinek2018l4,vaswani2019painless,paquette2020stochastic} consider local line search for every iteration in the stochastic setting. 
%
Most notably, \cite{schaul2013no} is perhaps the most closely related to our work, as they also consider a locally optimal learning rate based on a descent lemma for smooth functions. There are a few major differences: (1) Our work provide convergence analysis with exact rates that are optimal under certain assumptions, while their work only show an asymptotic convergence proof; (2) Our work propose an improved estimation of the optimal learning rate, and in Subsection \ref{subsec:norm estimation} we show empirically the advantage of our estimator; and (3) We also propose a model independent method to estimate the smoothness parameter, based on \cite{liu2023sophia}, while \cite{schaul2013no} rely on the structure of the model to estimate the smoothness.
In a follow-up to \cite{schaul2013no}, \cite{wu2018understanding} generalized their work to momentum SGD under the assumption of a convex objective. They also study cases in which the locally optimal learning rate have short-horizon bias and may not perform well on longer optimization tasks. In our work we empirically test our method on both short and long tasks (such as large text datasets) and show that it performs well across tasks.

\paragraph{Parameter-free optimization.}

There is a vast literature focusing on parameter-free algorithms, that can optimize a large class of functions with essentially no hyperparameter tuning. Many such works \citep{orabona2016coin,cutkosky2018black,foster2017parameter,luo2015achieving,jacobsen2022parameter} use online learning techniques to construct algorithms that achieve near-optimal convergence rate in the convex setting, these works are mostly of theoretical nature. Several works proposed more practical parameter-free algorithm with an empirical study of their methods \citep{orabona2014simultaneous,orabona2017training,kempka2019adaptive,chen2022better}. Recently \cite{ivgi2023dog} and \cite{defazio2023learning} have proposed parameter-free algorithms which are similar in flavor and focus on estimating the distance of the weights from the solution. While \cite{defazio2023learning} focus on the deterministic case, \cite{ivgi2023dog} prove convergence rate with high probability in the stochastic case with bounded noise. We note that all the above works consider the convex setting, while our work consider smooth and not-necessarily convex objectives.

\paragraph{Meta-learning techniques.} 

Several works consider different approaches to ``learn the learning rate'' in a dynamic method, while requiring either a few or no parameters to tune~\citep{baydin2017online,zhang2019lookahead,zhuang2020adabelief}.
\cite{baydin2017online} dynamically update the learning rate by calculating the gradient w.r.t the learning rate itself. 
\cite{zhang2019lookahead} proposes a method to ``lookahead'' several steps using a a different and presumably faster optimizer, and using this information to calculate an improved descent direction. \cite{zhuang2020adabelief} proposes a method of predict the next gradient based on previous gradients, and take a step only if the observed gradient is close to the prediction.










\section{Dynamic Step-size Based on Locally Optimal Step Size}
\label{sec:optimal learning rate}

In the following section we will first show how we derived the GLyDER stepsize scheduler which is based on finding an optimal stepsize at each iteration given a stochastic descent direction. We will next show how to efficiently estimate this optimal stepsize using several stochastic gradients, present the GLyDER stepsize algorithm and show a convergence result for it. 


\paragraph{Problem formulation.}
%
Throughout, we assume the following optimization setup for our technical derivations and theoretical analyses.
We consider an objective function $f:\reals^d\rightarrow\reals$ which is $L$-smooth,  We assume that at each point $\bx$ we have access to the gradient via a noisy oracle $\bg = \nabla f(\bx) + \bxi$ for some noise vector $\bxi$.  Our assumptions on the noise are $\E[\bxi \mid \bx]=0$ and $\E\left[\norm{\bxi}^2 \mid \bx \right]\leq \sigma$, i.e., it has zero mean and bounded variance conditioned on the current iteration. This oracle is standard in many machine learning applications where at each iteration a batch of data is sampled, and the gradient of the loss is calculated with respect to this batch.



\subsection{Deriving the GLyDER Stepsize}

Our starting point is the following descent lemma for $L$-smooth functions, which is  often used to give optimization guarantees (see \cite{bauschke10convex,beck2017first}):



\begin{theorem}[The descent lemma]\label{thm:descent lemma}
    Let $\bx'$ be defined by 
    $
        \bx' = \bx - \bd
    $
    for some arbitrary vector $\bd$.
    Then for any $L$-smooth function $f$ we have
    \begin{equation}\label{eq:descent lemma}
         f(\bx) - f(\bx') \geq \langle \nabla f(\bx), \bd\rangle - \frac{L}{2} \|\bd\|^2.
    \end{equation}
\end{theorem}


From the descent lemma we derive the optimal step size in terms of the best lower bound for the change in function value when moving in an arbitrary direction $\bd$, by considering a learning rate that maximizes the r.h.s of \eqref{eq:descent lemma}. This provides the following corollary, where by optimal we mean that it maximizes the change in function value when only adjusting the learning rate and keeping the descent direction constant.

\begin{corollary}\label{cor:descent}
Given a starting point $\bx$ and an arbitrary direction $\bd$, the optimal step size in the direction $\bd$ (according to the bound in \thmref{thm:descent lemma}) is: $\bx' = \bx - \frac{\langle \nabla f(\bx), \bd\rangle}{L \|\bd\|^2}\bd$, and the change in the function value is lower bounded by: $f(\bx) - f(\bx') \geq \frac{\inner{\nabla f(\bx), \bd}^2}{2L\norm{\bd}^2}$.
\end{corollary}

Consider running SGD for $t$ iterations, and at each iteration sampling a stochastic gradient $\bg_t$.
\corollaryref{cor:descent} states that given $\bg_t$, the optimal learning rate for this descent direction would be 
\begin{equation}\label{eq:greedy with inner prod}
\eta_t = \frac{\inner{\nabla f(\bx_t), \bg_t}}{L\norm{\bg_t}^2}~.
\end{equation}
Thus, given a sequence of stochastic gradients $\bg_1,\dots,\bg_T$, we can define a sequence of learning rates $\eta_1,\dots,\eta_T$ which are guaranteed to be locally optimal in the sense that at each step $t\in[T]$ we choose the optimal learning rate when taking a step in the direction $\bg_t$. Also, note that in the noiseless case, i.e. when $\bg_t = \nabla f(\bx_t)$, then the learning rate is constant and equal to $\frac{1}{L}$ which is known to give the optimal convergence rate for gradient descent using the descent lemma (see e.g. \cite{bubeck2015convex}).

 Our goal is to find a stationary point of $f$, that is, given $\epsilon >0$ finding $\bx\in\reals^d$ with $\norm{\nabla f(\bx)}^2 \leq \epsilon$. Note that it is not possible to provide stronger guarantees such as converging to a global or local minimum without further assumptions on the function (such as convexity, PL condition \cite{karimi2016linear} etc.)
In the following theorem we analyze the convergence rate of the learning rate scheduler from \eqref{eq:greedy with inner prod}:

\begin{theorem}\label{thm:optimal LR}
    Let $f:\reals^d\rightarrow \reals$ be an $L$-smooth function. Suppose we run SGD for $T$ iterations in the following way: We start at some arbitrary point $\bx_0$ with $|f(\bx_0)-f(\bx^*)| \leq R$ where $\bx^*$ is a global minimum of $f$. At each iteration $t\in[T]$ we sample $\bg_t = \nabla f(\bx_t) + \bxi_t$ where $\E[\bxi_t] = 0$ and $\E\left[\norm{\bxi_t}^2\right] \leq  \sigma^2$ for $\sigma >0$. We define $\bx_{t} = \bx_{t-1} - \frac{1}{L}\cdot \frac{\inner{\nabla f(\bx_t), \bg_t}}{\norm{\bg_t}^2}\bg_t$, then:
    \[
    \min_{t=1,\dots,T} \E\left[\norm{\nabla f(\bx_t)}^2\right] \leq \frac{2LR}{T} + \sigma\sqrt{\frac{2LR}{T}}
    \]
\end{theorem}

The full proof can be found in \appref{appen:proof of optimal LR}. The intuition behind the proof is pretty straightforward and requires iterative use of the descent lemma. The theorem shows that our learning rate scheduler have a couple of very useful properties:
\begin{itemize}[leftmargin=*]
    \item The convergence rate to reach a gradient of squared norm smaller than $\epsilon$ is $O\left(\ifrac{1}{\epsilon^2}\right)$. Due to previous lower bounds \citep{drori2020complexity}, this is known to be the optimal convergence rate in a noisy regime for smooth functions that are not necessarily convex.
    \item The learning rate doesn't require knowledge of either $\sigma$ or $R$, and is adaptive to both parameters.
    \item The convergence rate interpolates between the noisy regime ($\sigma >0$) and the noiseless regime ($\sigma = 0$).
\end{itemize}

To practically implement the learning rate described in \thmref{thm:optimal LR} we need to approximate the inner product $\inner{\nabla f(\bx_t), \bg_t}$ at each iteration. This approximation can be done by sampling a fresh mini-batch of stochastic gradients $\bh_t^1,\dots,\bh_t^n$ at each iteration and estimating the inner product:
\[
 \inner{\nabla f(\bx_t), \bg_t}  \approx \left\langle\frac{1}{n}\sum_{i=1}^n \bh_i^t, \bg_t\right\rangle ~.
\]

However, this approximation has one main caveat:
We cannot bound the term  $\inner{\nabla f(\bx_t), \bg_t}$ away from zero, hence it is not possible to use concentration bounds such as Hoefdding's inequality to achieve a good relative approximation where $n$ is constant for all iterations that depends only on the problem's parameter. 
To overcome this caveat, in the next subsection we provide a different analysis which estimate the squared norm of the gradient, instead of its inner product with the stochastic gradient.

\subsection{Approximation of the GLyDER Stepsize}
\label{subsec:practical implementation}

We begin this section with finding the locally optimal learning rate in expectation, which relies on a descent lemma in expectation. Recall that we consider an $L$-smooth function $f(\bx)$, and we are given at $\bx$ a stochastic gradient $\bg:=\nabla f(\bx) + \bxi$ where the noise is with zero mean, and $\E[\norm{\bxi}^2 \mid \bx] \leq \sigma^2$. Applying the smoothness condition for a descent direction $\eta(\nabla f(\bx) + \bxi)$ in expectation over the noise $\bxi$ we get:
\begin{align*}
\E\left[ f(\bx - \eta(\nabla f(\bx)+\bxi)) - f(\bx) \mid \bx\right] &\leq  \E\left[ -\eta\inner{\nabla f(\bx), \nabla f(\bx)+\bxi} + \eta^2L\frac{\| \nabla f(\bx)+\bxi\|^2}{2}\mid \bx\right]
\\
&= -\eta \|\nabla f(\bx)\|^2 + \eta^2L\frac{\|\nabla f(\bx)\|^2 + \sigma^2}{2}  
\end{align*}
minimizing the r.h.s over $\eta$ yields that the optimal descent rate in expectation is $\frac{1}{L}\cdot\frac{\norm{\nabla f(\bx)}^2}{\norm{\nabla f(\bx)}^2 + \sigma^2}$. As in the previous section, we can show that this learning rate scheduler achieves the optimal convergence rate:

\begin{theorem}\label{thm:expectation step size}
    Let $f:\reals^d\rightarrow \reals$ be an $L$-smooth function. Suppose we run SGD for $T$ iterations in the following way: We start at some arbitrary point $\bx_0$ with $|f(\bx_0)-f(\bx^*)| \leq R$ where $\bx^*$ is a global minimum of $f$. At each iteration $t\in[T]$ we sample $\bg_t = \nabla f(\bx_t) + \bxi_t$ where $\E[\bxi_t] = 0$ and $\E\left[\norm{\bxi_t}^2\right] \leq  \sigma^2$ for $\sigma >0$. We define $\bx_{t} = \bx_{t-1} - \frac{1}{L}\cdot \frac{\norm{\nabla f(\bx_t)}^2}{\norm{\nabla f(\bx_t)}^2 + \sigma^2}\bg_t$, then:
    \[
    \min_{t=1,\dots,T} \E\left[\norm{\nabla f(\bx_t)}^2\right] \leq \frac{2LR}{T} + \sigma\sqrt{\frac{2LR}{T}}~.
    \]
\end{theorem}
The full proof can be found in \appref{appen:proof of LR expectation}. Note that contrary to \thmref{thm:optimal LR}, here we do need to know $\sigma$ to define the learning rate, although we still don't need to know $R$. Note that also here, if we consider the noiseless case, then we get a constant learning rate of $\frac{1}{L}$.

The advantage of defining the locally optimal learning rate in this way, is that it can be efficiently estimated by sampling new gradients. Namely, by sampling stochastic gradients we can derive an unbiased estimator of the squared norm of the gradient. This is described by the following procedure:

Suppose that we sample i.i.d stochastic gradients $\bh_t^1,\dots,\bh_t^n$ at $\bx_t$, where $\bh_t^i = \nabla f(\bx_t) + \bxi_t^i$ with $\E[\bxi^t_i] = 0$ and bounded variance. Then $\frac{1}{n(n-1)}\sum_{i\neq j}\inner{\bh^i_t,\bh^j_t}$ is an unbiased estimator of the norm of the gradient. This is because:
\begin{align}\label{eq: unbiased norm estimator}
    \E\left[\frac{1}{n(n-1)}\sum_{i\neq j}\inner{\bh^i_t,\bh^j_t}\right] &= \frac{1}{n(n-1)}\E\left[\sum_{i\neq j}\inner{\nabla f(\bx_t) + \bxi_t^i,\nabla f(\bx_t) + \bxi_t^j}\right] \nonumber\\
    & = \frac{1}{n(n-1)}\E\left[\sum_{i\neq j} \inner{\nabla f(\bx_t), \nabla f(\bx_t)} + \inner{\nabla f(\bx_t), \bxi_t^i} + \inner{\nabla f(\bx_t), \bxi_t^j} + \inner{\bxi_t^i, \bxi_t^j}\right] \nonumber\\
    & = \frac{1}{n(n-1)}\sum_{i\neq j} \inner{\nabla f(\bx_t), \nabla f(\bx_t)} =  \norm{\nabla f(\bx_t)}^2~,
\end{align}
where we used that the noise is sampled i.i.d with zero mean. Note that to estimate the denominator of the learning rate, we can either estimate $\norm{\nabla f(\bx_t)}^2 $ and $\sigma^2$ separately, or we could just use  $\frac{1}{n^2}\sum_{i,j =1}^n\inner{\bh_i^t,\bh_j^t}$ which is an unbiased estimator of $\norm{\nabla f(\bx_t)}^2  + \ifrac{\sigma^2}{n}$.

One of the advantages of this estimator is 
that it aligns with the goal of the algorithm, to minimize the squared norm of the gradient. In more details, suppose we want to achieve a squared gradient norm smaller than some $\epsilon > 0$, then either the norm of the gradient is larger than $\epsilon$, and we can estimate it using a mini-batch of size $O\left(\ifrac{1}{\epsilon^2}\right)$, or it is smaller than $\epsilon$, which means our algorithm converged to an $\epsilon$-stationary point.


Since this estimation reduces the variance of the noise by a factor of $n$ (i.e. $\ifrac{\sigma^2}{n}$ instead of $\sigma^2$), it is natural to also sample a mini-batch of size $n$ to determine the descent direction and reduce the variance by the same factor. The reason for sampling two different sets of stochastic gradients is only for theoretical reasons, to avoid the dependence between the learning rate calculation and the descent direction. In our experimental results we use the same set of gradients for both tasks.

\subsection{The GLyDER stepsize Algorithm }

We now present in \algref{alg:greedy constant L} the GLyDER stepsize algorithm and provide convergence guarantees for it.
The input for the algorithm is a starting point $\bx_0$, the smoothness of the objective function $L$ and the number of stochastic gradient $n$ that are sampled at each iteration. The algorithm at each iteration calculates unbiased estimators $\mu_t \sim \norm{\nabla f(\bx_t)}^2$ and $\gamma_t \sim \norm{\nabla f(\bx_t)}^2 + \sigma^2 /n$, where $\nabla f(\bx)$ is the gradient (i.e. non-stochastic) of $f$ at $\bx$. Finally, the algorithm performs an update in the direction of the sum of the stochastic gradients, and with a step size $\frac{1}{L}\cdot\frac{\mu_t}{\gamma_t}$. 



\begin{wrapfigure}[13]{L}{0.5\textwidth}
\begin{algorithm}[H]
 \textbf{Input:} $\bx_0,~ n, ~L$
 
 \For{$t=1,2,\dots,T$}{
  \textbf{Sample} stochastic gradients $\bg_t^1,\dots,\bg_t^n, \bh_t^1,\dots, \bh_t^n$\\
  \textbf{Set:} \\
  ~~~~ $\mu_t:= \frac{1}{n(n-1)}\sum_{i\neq j} \inner{\bh_t^i,\bh_t^j}$ \\
  ~~~~ $\gamma_t:= \frac{1}{n^2}\sum_{i,j=1}^m \inner{\bh_t^i,\bh_t^j}$ \\
  ~~~~ $\bg_t = \sum_{i=1}^n\bg_t^i$
  
  
  \textbf{Update} $\bx_t = \bx_{t-1} - \frac{1}{L}\cdot\frac{\mu_t}{\gamma_t}\bg_t$
  }
 \caption{Theoretical GLyDER learning rate}
 \label{alg:greedy constant L}
\end{algorithm}
\end{wrapfigure}

The following theorem shows the convergence of \algref{alg:greedy constant L}:

\begin{theorem}\label{thm:sample to approx step size}
    Let $f:\reals^d\rightarrow \reals$ be an $L$-smooth and $M$-Lipschitz function and let $\epsilon > 0$. Suppose $\bx_0$ is such that $|f(\bx_0)-f(\bx^*)| \leq R$ where $\bx^*$ is a global minimum of $f$. 
    Also assume that the norm of the noise vectors are globally bounded by $G >0$.
    Suppose that we run \algref{alg:greedy constant L} with  $n \geq \ifrac{20\sigma^2}{\epsilon}$, then there exists a universal constant $c>0$ such that w.p $> 1- dT\exp\left(-\frac{\epsilon^2 n c}{\sigma^2M^2G^2}\right)$, we get:
    \[
    \min_{t=1,\dots,T}\E\left[\norm{\nabla f(\bx_t)}^2\right] \leq \frac{2LR}{T} + \frac{\sigma}{\sqrt{n}}\sqrt{\frac{2LR}{T}}~,
    \]
    where the high probability is w.r.t the noise of the stochastic gradients $\bh_t^i$'s, and the expectations is w.r.t the stochastic gradients $\bg_t^i$'s.
    In particular, by choosing  $n=\Omega\left(\ifrac{\sigma^2M^2G^2\log(dT)}{\epsilon^2}\right)$, then w.p $ > \Omega(1)$ the oracle complexity to achieve 
    $
        \min_{t=1,\dots,T}\E\left[\norm{\nabla f(\bx_t)}^2\right] \leq \epsilon
    $
    is $O\left(\ifrac{\sigma^2M^2 G^2L^2R^2\log(dT)}{\epsilon^3}\right)$.
\end{theorem}

The full proof can be found in \appref{appen: proof of sample to approx step size}. 
The reason that we use two sets of stochastic gradients is to avoid the dependence for the theoretical analysis. In \secref{sec:practical implementation} we provide the practical algorithm which uses the same set of gradients for both calculations, and our experiments are done on this practical algorithm. Also note that the algorithm converges to a gradient with squared norm smaller than $\epsilon$ in $O\left(\ifrac{1}{\epsilon^3}\right)$ oracle calls. This is not the optimal convergence rate which we achieved in \thmref{thm:optimal LR} and \thmref{thm:expectation step size}. The reason is that here we don't assume to known $\sigma$, and estimating it up to an error of $O(\epsilon)$ requires $\Omega\left(\ifrac{1}{\epsilon^2}\right)$ oracle calls. We are not aware of another method which achieves the optimal learning rate without knowledge of $\sigma$, and with only logarithmic dependence on the input dimension.
\section{Practical Implementation}\label{sec:practical implementation}

In the previous section we  provided theoretical justification for GLyDER. In this section we discuss several practical considerations and heuristics which improve its empirical performance.


\subsection{Smoothness Estimation}\label{subsec:smoothness estimation}

Our learning rate scheduler requires knowing the smoothness of the function. In practice, not only that in general there is no prior knowledge of this term, the smoothness may also not be bounded. This is in fact the case for modern neural network which are known to be non-smooth and non-convex. 
One naive method to approximate the smoothness is to assume it is globally bounded, and to run a hyperparameter sweep with a goal of finding an upper bound which will work for the entire optimization process.

A different approach is to use an adaptive method to approximate the smoothness at every iteration. Here we provide two methods for such an approximation, which are inspired by similar methods from \cite{liu2023sophia}. For this subsection we assume a standard supervised setting: We have a loss function $\ell(\cdot,\cdot)$, a model $N(\btheta,\bx)$ (e.g a neural network) parameterized by $\btheta$ with input data $\bx$. For a batch of samples $(\bx_1,y_1),\dots,(\bx_m,y_m)$ and parameters $\btheta$ we can define the empirical loss as $f(\btheta):=\frac{1}{m}\sum_{i=1}^m \ell_{\text{ce}}(N(\btheta,\bx_i), y_i)$\footnote{Only in this subsection, we slightly abuse notation and consider the parameters of the objective function as $\btheta$ instead of $\bx$, since}. In this setting sampling a stochastic gradient means sampling a batch of inputs.


\textbf{Option 1: Projection to a $1$-dimensional function ($1$-d projection).} For a function multivariate $f:\reals^d\rightarrow \reals$, if we pick at some point $\btheta_t$ a descent direction $\bg_t$ (i.e. by sampling a batch of inputs), finding the best learning $\eta_t$ for this direction boils down to a one-dimensional function. Thus, the smoothness of our function is defined by the second derivative of this $1$-dimensional function. That is, we can calculate: $L_t := \bg_t^\top \nabla^2 f(\btheta_t) \bg_t\in \reals$
as the smoothness bound for iteration $t$. The second derivative is calculated w.r.t the same batch of inputs that was used to calculate the gradient. Note that although this term contains the Hessian of $f$, it requires only calculating the projection of the Hessian in the direction $\bg_t$. This projection can be calculated in time $O(d)$, which is similar (up to constant factors) to the computational cost of calculating the gradient. In theory, the smoothness of the function could significantly change across this line, and line search could be used to gain a better bound. For practical considerations we do not perform line search, and take $L_t$ as the local smoothness bound.

\textbf{Option 2: A simplified version of the Gauss-Newton-Bartlett (GNB) estimator.} In \cite{liu2023sophia} (Section 2.3, Option 2) the authors consider a multi-class classification problem. 
They consider the cross-entropy loss, where $N$ is some model (e.g. a neural network) parameterized by $\btheta$ with input data $\bx$, and its output $N(\btheta,\bx)\in \reals^V$ are the logits, where $V$ is the number of possible classes, and $y\in\{1,\dots,V\}$ is the output class. This represents the loss function for a multi-class classification problem with $V$ classes on a single sample.

For a batch of samples $(\bx_1,y_1),\dots,(\bx_m,y_m)$ and parameters $\btheta$ we can define the empirical loss as $L(\btheta):=\frac{1}{m}\sum_{i=1}^m \ell_{\text{ce}}(N(\btheta,\bx_i), y_i)$. We consider the estimator: $L_t := \max\left\{\nabla_{\btheta}L(\btheta) \odot \nabla_{\btheta}L(\btheta)  \right\}$,
where $\odot$ is the Hadamard-product, and the $\max$ function is over the coordinates \footnote{We note that in \cite{liu2023sophia}, the authors consider a more complicated estimator. Namely, given $t=N(\btheta,\bx_i)\in\reals^V$ they consider the categorical distribution $\text{Cat}(t)$ (a categorical distribution over the vector of logits $t$), they sample a label $\hat{y}_i$ from this distribution and calculate the empirical loss using these labels. We use the simpler estimator, which is also easier to calculate, without sampling labels from the distribution $\text{Cat}(t)$. In our empirical evaluation this estimator seems to work well in most tasks, the reason might be that our goal is much simpler -- finding the largest value in the diagonal of the Hessian, instead of estimating the entire diagonal in \cite{liu2023sophia}}.

We note that both options are easy to calculate. The advantage of the first method is that the reasoning behind it is clearer and more straightforward, although it does require calculating a projected second derivative, which takes approximately the same time as calculating the gradient. The second option is faster to calculate, as it requires almost no extra calculations, but is viable in theory only for the cross-entropy loss. This method can be considered as an easily calculated heuristic.


\subsection{Exponential Averaging and Norm Estimation}\label{subsec:norm estimation}

Our learning rate scheduler depends on estimating the norm of the real gradient using stochastic gradients. By only assuming that the noise of the stochastic gradients are independent and with zero mean, we constructed an unbiased estimator in \eqref{eq: unbiased norm estimator}. But having an unbiased estimator is usually not good enough, we would also like our estimator to have low-variance. In the following theorem we calculate the variance of this estimator in the general case and under an additional assumption that the noise is Gaussian.

\begin{theorem}\label{thm:variance of the norm}
    Let $n, d\in\mathbb{N}$ with $n,d>1$. Assume for each $i\in[n]$ we are given independent noisy estimate of the gradient, That is, $\bg_{i} = \nabla + \bxi_{i}$ for some $\nabla \in \reals^d$, where $\E[\bxi_{i}] = 0$, $\E[\norm{\bxi_{i}}^2] = \sigma^2$, and $\bg_i, \nabla, \bxi_i\in\reals^d$. Define $\mu = \frac{1}{n(n-1)}\sum_{i\neq j}\inner{\bg_{i}, \bg_{j}}$, then we have $\E[\mu] = \norm{\nabla}^2$ and $\var(\mu)  \leq \frac{4\norm{\nabla}^2\sigma^2}{n} + \frac{\sigma^4}{n(n-1)} $. In particular, for $\bxi_i\sim\Ncal\left(0,\frac{\sigma^2}{d}I_d\right)$ we have     $\var(\mu) = \frac{4\norm{\nabla}^2\sigma^2}{nd} + \frac{\sigma^4}{dn(n-1)}$.
\end{theorem}

\begin{wrapfigure}[19]{t}{0.4\textwidth}
\centering
\includegraphics[width=0.4\textwidth]{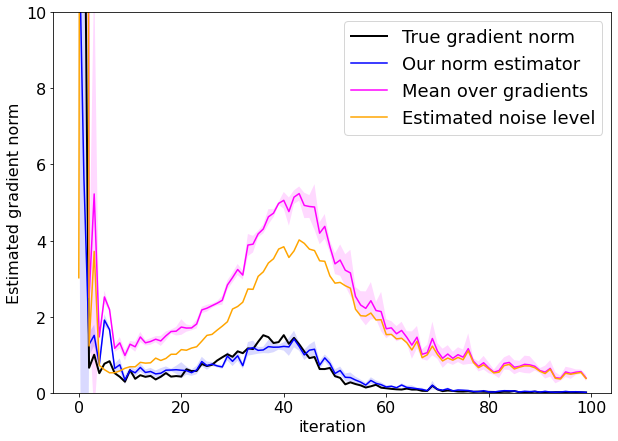}
\caption{\footnotesize Comparison of our norm estimator (blue) with the norm over the mean of the stochastic gradients (magenta), and the full gradient (black). The estimated noise level (orange) is the subtraction between the two. 
\label{fig:norm estimation}}
\end{wrapfigure}

The proof can be found in \appref{appen:proofs from practical}. Note that without further assumptions on the noise we can only upper bound the variance, as it will depend on the inner products between the noise vectors $\bxi_i$'s, and on the inner product between the noise vectors, and $\Delta$ (which represent the gradient).
For a Gaussian distribution, which is spherically symmetric and each coordinate is distributed i.i.d these inner products can be calculated and we can give an exact expression of the variance. This expression is smaller than our upper bound by a factor of $d$ (the input dimension).

It is interesting to note that the variance exhibits two phases during training depending on the norm of the gradient: (1) When $\norm{\nabla f(\bx)}^2$ is large, the variance decreases by a factor of $O\left(\ifrac{1}{n}\right)$; (2) When $\norm{\nabla f(\bx)}^2$ is very close to zero, the variance decreases by a larger factor of $O\left(\ifrac{1}{n^2}\right)$, but the effect of the noise $\sigma$ is also larger.
The variance reduction in the second phase is achieved thanks to the fact that although we only receive $n$ stochastic gradients, using all the inner products yields $O(n^2)$ (unbiased) estimators of the norm. Although these estimators are not independent, they still allow for variance reduction, at least when the norm is small.

If \figref{fig:norm estimation} we plot our norm estimator, compared to the norm of the mean over the gradients (which is used as the norm estimator in \cite{schaul2013no}). We also plot the true gradient, i.e. the gradient w.r.t the entire dataset. Note that our estimator closely follow the real gradient, while using the mean estimator is biased. The experiment is done on the CIFAR100 dataset, where each stochastic gradient is drawn independently $5$ times, showing the standard deviation of the estimators. For full experimental details see \appref{appen: exp details}.

Our bounds show that the noise has a very significant effect on the performance of our estimator. A method to reduce the variance which is commonly used in modern optimization algorithms (e.g. Adam \cite{kingma2014adam}) is to add exponential averaging. 
Namely, consider the newly estimated learning rate $\eta_t$ at iteration $t$ which is calculated in \thmref{thm:sample to approx step size}. For $\beta\in (0,1)$ use exponential averaging to get the learning rates: $\hat{\eta}_t = (1 - \beta) \eta_t + \beta\hat{\eta}_{t-1}$,
where $\hat{\eta}_t$ is the learning rate after exponential averaging. In all of our experiments, we used $\beta = 0.999$ as the coefficient for exponential averaging, which achieved good performance throughout all the tested datasets, algorithms and models. We note that also other optimization techniques uses exponential averaging (e.g Adam \cite{kingma2014adam}), with a constant coefficient that achieves good performance across different tasks, thus although there is an added hyper-parameter, in practice we do not tune it in any of our experiments.

\begin{wrapfigure}[19]{L}{0.455\textwidth}

\begin{algorithm}[H]\label{alg:greedy vary L}
 \textbf{Input:} $\bx_0,~ n, ~\eta_0, \beta$
 
 \For{$t=1,2,\dots,T$}{
  \textbf{Sample} stochastic gradients $\bg_t^1,\dots,\bg_t^n$\\
  \textbf{Set:} \\
  ~~~~ $\mu_t:= \left\|\sum_{i=1}^n\bg^i_t\right\|^2 - \sum_{i=1}^n\left\|\bg^i_t\right\|^2$ \\
  ~~~~ $\gamma_t:= \left\|\sum_{i=1}^n\bg^i_t\right\|^2 $ \\
  ~~~~ $\bg_t = \sum_{i=1}^n\bg_t^i$
  
  \If {$\gamma_t = 0$ or $\mu_t \leq 0$}{Set $\frac{\mu_t}{\gamma_t}:= 0$}
  
  Estimate $L_t$ using either option $1$ or $2$ from Subsection \ref{subsec:smoothness estimation}
  
  \textbf{Set:} $\eta_t := (1-\beta)\eta_{t-1} + \beta\cdot\frac{1}{L_t}\cdot \frac{\mu_t}{\gamma_t}$
  
  \textbf{Update} $\bx_t = \bx_{t-1} - \eta_t\bg_t$
  }
 \caption{Practical GLyDER learning rate scheduler}
\end{algorithm}
\end{wrapfigure}



\subsection{A Practical Implementation of GLyDER}

To summarize all the practical considerations discussed in this section, in \algref{alg:greedy vary L} we present the full version of our learning rate scheduler. Note that we use the same set of stochastic gradients to calculate both the descent direction and learning rate at each iteration, in contrary to the more theoretical \algref{alg:greedy constant L}.

There are two more minor considerations which we take into account when implementing our learning rate scheduler, which we discuss more extensively in \appref{appen:additional practical details}. In a nutshell: (1) We use an efficient implementation of the unbiased norm estimator, requiring only $O(n)$ operations instead of $O(n^2)$ when considering all the inner products. (2) We describe how to use parallel computational processors such as TPUs \citep{jouppi2017datacenter} to simulate several stochastic gradients without the need of re-sampling.

Finally, we can use GLyDER as a stepsize scheduler wrapper around any optimization algorithm, beyond vanilla SGD. We provide the full algorithm and additional details in \appref{appen:greedy extension}. In short, the only change from the original algorithm is when estimating the smoothness using option 1 from Subsection \ref{subsec:smoothness estimation} we project onto the descent direction provided by the optimizer, instead of onto the direction if the gradient.

\section{Experiments}\label{sec:experiments}

We compared our "GLyDER" stepsize scheduler to several standard manually tuned schedulers on a range of different machine learning tasks. In our experiments we varied both the models and the training algorithms to illustrate the effectiveness of our scheduler across many different scenarios. Our experiments are done using the init2winit framework \citep{init2winit2021github} which is based on JAX \citep{jax2018github}.

\vspace{-2pt}
\subsection{Experimental Details}
\paragraph{Methodology and training algorithms.} We compared our GLyDER stepsize scheduler to three standard schedulers: constant, cosine and reversed ``squashed'' square root (rsqrt) which is defined as $\eta_t = \eta_0\cdot \frac{\sqrt{s}}{\sqrt{t + s}}$ where $\eta_0$ is the initial learning rate and $s$ is the number of "squash" steps. We also compared our scheduler using three different training algorithms: SGD, momentum SGD and Adam \citep{kingma2014adam}. 

On each scheduler, dataset and training algorithm we performed a hyper-parameter search over a large parameter space. After choosing the best-performing hyperparameters we conducted five new experiments with different random seeds, and report the mean and standard deviation of the results. For the full hyper-parameter range of the scheduler and optimizers see \appref{appen: exp details}.
Training was done using a TPU \cite{jouppi2017datacenter} containing $8$ chips. In practice, for the GLyDER scheduler it means that at \algref{alg:greedy vary L} we received at each iteration $n=8$ stochastic gradients which were used to estimate the GLyDER stepsize, while those gradients were also used to calculate the descent direction.

\paragraph{Datasets.} We conducted the experiments on  datasets from the fields of vision, NLP and recommendation systems:
\begin{enumerate}[leftmargin=*]
    \item CIFAR10 and CIFAR100 \citep{krizhevsky2009learning}, evaluated by error percentage on the test set. 
    \item Imagenet \citep{russakovsky2015imagenet}, evaluated by test error on the test set. We trained on the entire dataset (containing $\sim 1M$ samples), and all the images are scaled to size $224 \times 224$.
    \item WikiText-2 \citep{merity2016pointer} which contains over $2M$ words extracted from Wikipedia, and is a standard language modeling benchmark. We evaluated on a left out validation set, and report the validation error, as well as the perplexity (in the appendix).
    \item Criteo 1TB\footnote{https://www.kaggle.com/c/criteo-display-ad-challenge} which contains feature values and click feedback for millions of display ads. This is a recommendation task for click-through rate prediction. We report the area under the curve (AUC) metric.
\end{enumerate}
For each dataset we used the default model from the init2winit framework. Notably, we compared the GLyDER scheduler across different models solving different types of tasks, including: Wide-ResNet \citep{zagoruyko2016wide}, ResNet50 \citep{he2016deep}, DLRM \citep{naumov2019deep} and LSTM \citep{wiseman2016sequence}. For full details about the trained models see \appref{appen: exp details}.

\subsection{Results}
In Table \ref{tab:results momentum} we compare the performance of the GLyDER scheduler using both options from Subsection \ref{subsec:smoothness estimation} to estimate the smoothness, i.e. either using the smoothness of the projection to a $1$-dimensional function, or the variation of the Gauss-Newton-Bartlett (GNB) estimator. All the experiments in Table \ref{tab:results momentum} are done with the momentum SGD optimizer. From the results it can be seen that the GLyDER stepsize scheduler, with either one of the smoothness estimator, matches the best performing tuned scheduler up to an error of less than $1\%$ across all the datasets. We performed similar experiments on vanilla SGD and Adam, showing that the GLyDER stepsize also generalize across different algorithms, for the full results see \appref{appen: additional experiments}.

In \figref{fig:stepsize error rate} we compare the GLyDER scheduler with $1$-dimensional projection to the other schedulers for the CIFAR10 and Imagenet datasets. The figure depicts an interesting advantage of the GLyDER stepsize, although it achieves similar performance to the other best performing schedulers, it converges much faster to that solution, around two thirds of the number of steps it takes to the other schedulers. We emphasize that it is possible to run the other schedulers for less training steps, but it reduces their performance, specifically the cosine scheduler which matches the performance of the GLyDER scheduler requires a certain number of steps to reach its optimal solution. Also, note the although the GLyDER stepsize is very small, it is certainly not constant. In fact, it exhibits a sudden drop in the stepsize during its run. It is evident that the GLyDER schedulers is adaptive to the task, and is effected by the gradients which can be seen as an advantage, and it would be interesting to further study this sudden stepsize drop behavior in future works.

\begin{table}[t]
\begin{center}
\begin{tabular}{|l|lllll|}
\hline
                              & \multicolumn{1}{l|}{\textbf{GLyDER + $1$-- d proj}} & \multicolumn{1}{l|}{\textbf{GLyDER + GNB}} & \multicolumn{1}{l|}{\textbf{Constant}} & \multicolumn{1}{l|}{\textbf{Cosine}}  & \textbf{rsqrt}  
                           \\   \hline
\textbf{CIFAR10 $\downarrow$}              & \multicolumn{1}{l|}{$3.0\% \pm 0.1$}             & \multicolumn{1}{l|}{$4.6\% \pm 0.2$}         & \multicolumn{1}{l|}{$4.7\% \pm 0.07$}    & \multicolumn{1}{l|}{$3.0\% \pm 0.1$}    & $4.3 \% \pm 0.1$ \\ \hline
\textbf{CIFAR100 $\downarrow$}             & \multicolumn{1}{l|}{$19.7 \%\pm 0.3$}            & \multicolumn{1}{l|}{$22.8 \%\pm 1.3$}        & \multicolumn{1}{l|}{$22.6 \%\pm 0.8$}    & \multicolumn{1}{l|}{$19.2 \%\pm 0.2$}   & $21.3\% \pm 0.7$               \\ \hline
\textbf{Imagenet $\downarrow$}             & \multicolumn{1}{l|}{$24.6\% \pm 0.1$}            & \multicolumn{1}{l|}{$25.6\% \pm 1.1$}        & \multicolumn{1}{l|}{$32.3 \pm 0.1$}    & \multicolumn{1}{l|}{$23.7\% \pm 0.08$}  & $28.7 \%\pm 0.1$               \\ \hline
\textbf{WikiText-2 $\downarrow$}        & \multicolumn{1}{l|}{$78.3 \%\pm 1.2$}            & \multicolumn{1}{l|}{$76.7\% \pm 0.03$}       & \multicolumn{1}{l|}{$75.8 \%\pm 0.0$}    & \multicolumn{1}{l|}{$75.9\% \pm 0.04$}  & $75.9\% \pm 0.0$               \\ \hline

\textbf{Criteo $\uparrow$}               & \multicolumn{1}{l|}{$0.78 \pm 0.003$}          & \multicolumn{1}{l|}{$0.69 \pm 0.006$}      & \multicolumn{1}{l|}{$0.78 \pm 0.001$}  & \multicolumn{1}{l|}{$0.75 \pm 0.004$}                 &            \multicolumn{1}{l|}{$0.7 \pm 0.003$}                    \\ \hline
\end{tabular}
\caption{Comparison of the GLyDER stepsize scheduler with the two options to estimate the smoothness trained using momentum SGD. 
For all the datasets except Criteo we report the top-$1$ error percentage on the test set (i.e. smaller is better), and for Criteo we report the AUC metric (larger is better). 
}\label{tab:results momentum}
\end{center}
\end{table}

\begin{figure}[t]
     \centering
     \begin{subfigure}[b]{0.45\textwidth}
         \centering
         \includegraphics[scale=0.3]{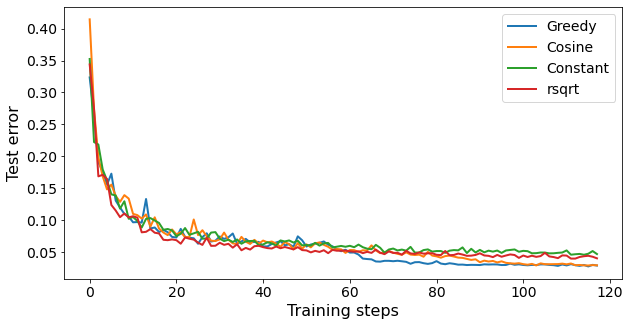}
         \caption{CIFAR10 -- error rate}
     \end{subfigure}
     \hfill
     \begin{subfigure}[b]{0.45\textwidth}
         \centering
         \includegraphics[scale=0.3]{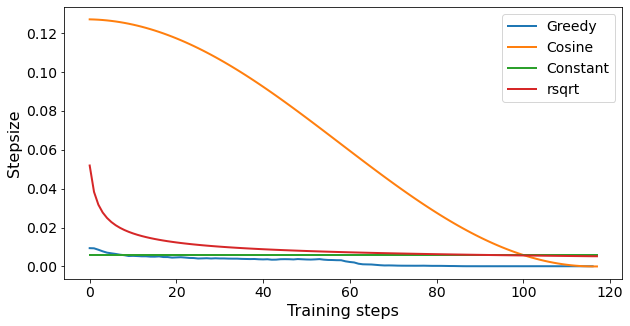}
         \caption{CIFAR10 -- stepsizes}
     \end{subfigure}
     \hfill
     \\
      \begin{subfigure}[b]{0.45\textwidth}
     \centering
     \includegraphics[scale=0.3]{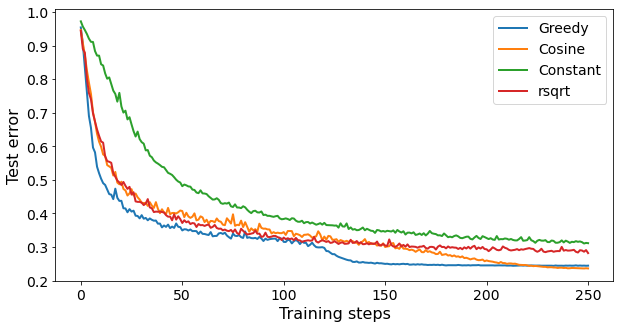}
     \caption{Imagenet -- error rate}
    \end{subfigure}
    \hfill
    \begin{subfigure}[b]{0.45\textwidth}
     \centering
     \includegraphics[scale=0.3]{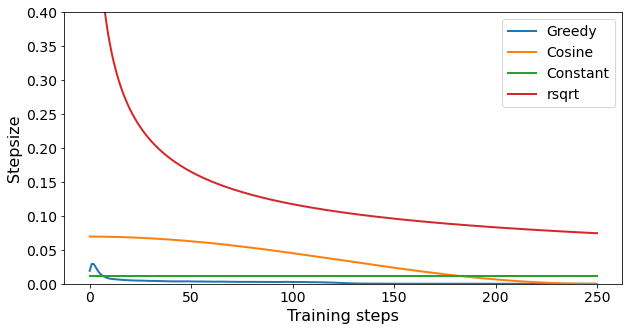}
     \caption{Imagenet -- stepsizes}
    \end{subfigure}
        \caption{Comparison of the error rate and stepsize for the CIFAR10 and Imagenet datasets between the GLyDER scheduler and other manually tuned schedulers.}
        \label{fig:stepsize error rate}
\end{figure}
    



\section{Discussion and Future Work}
We introduced GLyDER, a stepsize scheduler that determines optimal step sizes in the presence of stochastic descent directions. It attains optimal rate in the smooth (not necessarily convex) setting. We also introduce heuristic techniques to enhance its performance. Experimental results show that GLyDER performs on par with manually fine-tuned schedulers.

There are several future research directions which we think could be interesting to study. First, to gain theoretical understanding on how GLyDER performs beyond SGD, e.g. when adding momentum, or utilizing per-parameter stepsizes. Second, extending the theoretical understanding of smoothness estimation, not confined solely to the scope of the GLyDER scheduler, presents a promising area for investigation.. Finally, it would be interesting to further test and improve the GLyDER scheduler to other types of tasks, such as fine-tuning, self-supervised learning and different network architectures.

\bibliographystyle{authordate1}
\bibliography{bib}{}

\begin{thebibliography}{}

\bibitem[\protect\citename{Abadi {\em et~al.}, }2016]{abadi2016tensorflow}
Abadi, Mart{\'\i}n, Barham, Paul, Chen, Jianmin, Chen, Zhifeng, Davis, Andy, Dean, Jeffrey, Devin, Matthieu, Ghemawat, Sanjay, Irving, Geoffrey, Isard, Michael, {\em et~al.} 2016.
\newblock $\{$TensorFlow$\}$: a system for $\{$Large-Scale$\}$ machine learning.
\newblock {\em Pages  265--283 of:} {\em 12th USENIX symposium on operating systems design and implementation (OSDI 16)}.

\bibitem[\protect\citename{Anil {\em et~al.}, }2020]{anil2020scalable}
Anil, Rohan, Gupta, Vineet, Koren, Tomer, Regan, Kevin, \& Singer, Yoram. 2020.
\newblock Scalable second order optimization for deep learning.
\newblock {\em arXiv preprint arXiv:2002.09018}.

\bibitem[\protect\citename{Attia \& Koren, }2023]{pmlr-v202-attia23a}
Attia, Amit, \& Koren, Tomer. 2023.
\newblock {SGD} with {A}da{G}rad Stepsizes: Full Adaptivity with High Probability to Unknown Parameters, Unbounded Gradients and Affine Variance.
\newblock {\em Pages  1147--1171 of:} {\em Proceedings of the 40th International Conference on Machine Learning}.
\newblock Proceedings of Machine Learning Research, vol. 202.
\newblock PMLR.

\bibitem[\protect\citename{Bauschke \& Combettes, }2011]{bauschke10convex}
Bauschke, HH, \& Combettes, PL. 2011.
\newblock Convex Analysis and Monotone Operator Theory in Hilbert Spaces, 2011.
\newblock {\em CMS books in mathematics). DOI}, {\bf 10}, 978--1.

\bibitem[\protect\citename{Baydin {\em et~al.}, }2017]{baydin2017online}
Baydin, Atilim~Gunes, Cornish, Robert, Rubio, David~Martinez, Schmidt, Mark, \& Wood, Frank. 2017.
\newblock Online learning rate adaptation with hypergradient descent.
\newblock {\em arXiv preprint arXiv:1703.04782}.

\bibitem[\protect\citename{Beck, }2017]{beck2017first}
Beck, Amir. 2017.
\newblock {\em First-order methods in optimization}.
\newblock SIAM.

\bibitem[\protect\citename{Bradbury {\em et~al.}, }2018]{jax2018github}
Bradbury, James, Frostig, Roy, Hawkins, Peter, Johnson, Matthew~James, Leary, Chris, Maclaurin, Dougal, Necula, George, Paszke, Adam, Vander{P}las, Jake, Wanderman-{M}ilne, Skye, \& Zhang, Qiao. 2018.
\newblock {JAX}: composable transformations of {P}ython+{N}um{P}y programs.

\bibitem[\protect\citename{Bubeck {\em et~al.}, }2015]{bubeck2015convex}
Bubeck, S{\'e}bastien, {\em et~al.} 2015.
\newblock Convex optimization: Algorithms and complexity.
\newblock {\em Foundations and Trends{\textregistered} in Machine Learning}, {\bf 8}(3-4), 231--357.

\bibitem[\protect\citename{Chen {\em et~al.}, }2022]{chen2022better}
Chen, Keyi, Langford, John, \& Orabona, Francesco. 2022.
\newblock Better parameter-free stochastic optimization with ODE updates for coin-betting.
\newblock {\em Pages  6239--6247 of:} {\em Proceedings of the AAAI Conference on Artificial Intelligence},  vol. 36.

\bibitem[\protect\citename{Cutkosky \& Orabona, }2018]{cutkosky2018black}
Cutkosky, Ashok, \& Orabona, Francesco. 2018.
\newblock Black-box reductions for parameter-free online learning in banach spaces.
\newblock {\em Pages  1493--1529 of:} {\em Conference On Learning Theory}.
\newblock PMLR.

\bibitem[\protect\citename{Defazio \& Mishchenko, }2023]{defazio2023learning}
Defazio, Aaron, \& Mishchenko, Konstantin. 2023.
\newblock Learning-rate-free learning by D-adaptation.
\newblock {\em arXiv preprint arXiv:2301.07733}.

\bibitem[\protect\citename{Drori \& Shamir, }2020]{drori2020complexity}
Drori, Yoel, \& Shamir, Ohad. 2020.
\newblock The complexity of finding stationary points with stochastic gradient descent.
\newblock {\em Pages  2658--2667 of:} {\em International Conference on Machine Learning}.
\newblock PMLR.

\bibitem[\protect\citename{Duchi {\em et~al.}, }2011]{duchi2011adaptive}
Duchi, John, Hazan, Elad, \& Singer, Yoram. 2011.
\newblock Adaptive subgradient methods for online learning and stochastic optimization.
\newblock {\em Journal of machine learning research}, {\bf 12}(7).

\bibitem[\protect\citename{Foster {\em et~al.}, }2017]{foster2017parameter}
Foster, Dylan~J, Kale, Satyen, Mohri, Mehryar, \& Sridharan, Karthik. 2017.
\newblock Parameter-free online learning via model selection.
\newblock {\em Advances in Neural Information Processing Systems}, {\bf 30}.

\bibitem[\protect\citename{Ghadimi \& Lan, }2013]{ghadimi2013stochastic}
Ghadimi, Saeed, \& Lan, Guanghui. 2013.
\newblock Stochastic first-and zeroth-order methods for nonconvex stochastic programming.
\newblock {\em SIAM Journal on Optimization}, {\bf 23}(4), 2341--2368.

\bibitem[\protect\citename{Gilmer {\em et~al.}, }2021]{init2winit2021github}
Gilmer, Justin~M., Dahl, George~E., \& Nado, Zachary. 2021.
\newblock {init2winit}: a JAX codebase for initialization, optimization, and tuning research.

\bibitem[\protect\citename{Gupta {\em et~al.}, }2018]{gupta2018shampoo}
Gupta, Vineet, Koren, Tomer, \& Singer, Yoram. 2018.
\newblock Shampoo: Preconditioned stochastic tensor optimization.
\newblock {\em Pages  1842--1850 of:} {\em International Conference on Machine Learning}.
\newblock PMLR.

\bibitem[\protect\citename{He {\em et~al.}, }2016]{he2016deep}
He, Kaiming, Zhang, Xiangyu, Ren, Shaoqing, \& Sun, Jian. 2016.
\newblock Deep residual learning for image recognition.
\newblock {\em Pages  770--778 of:} {\em Proceedings of the IEEE conference on computer vision and pattern recognition}.

\bibitem[\protect\citename{Ivgi {\em et~al.}, }2023]{ivgi2023dog}
Ivgi, Maor, Hinder, Oliver, \& Carmon, Yair. 2023.
\newblock DoG is SGD's Best Friend: A Parameter-Free Dynamic Step Size Schedule.
\newblock {\em arXiv preprint arXiv:2302.12022}.

\bibitem[\protect\citename{Jacobsen \& Cutkosky, }2022]{jacobsen2022parameter}
Jacobsen, Andrew, \& Cutkosky, Ashok. 2022.
\newblock Parameter-free mirror descent.
\newblock {\em Pages  4160--4211 of:} {\em Conference on Learning Theory}.
\newblock PMLR.

\bibitem[\protect\citename{Jouppi {\em et~al.}, }2017]{jouppi2017datacenter}
Jouppi, Norman~P, Young, Cliff, Patil, Nishant, Patterson, David, Agrawal, Gaurav, Bajwa, Raminder, Bates, Sarah, Bhatia, Suresh, Boden, Nan, Borchers, Al, {\em et~al.} 2017.
\newblock In-datacenter performance analysis of a tensor processing unit.
\newblock {\em Pages  1--12 of:} {\em Proceedings of the 44th annual international symposium on computer architecture}.

\bibitem[\protect\citename{Karimi {\em et~al.}, }2016]{karimi2016linear}
Karimi, Hamed, Nutini, Julie, \& Schmidt, Mark. 2016.
\newblock Linear convergence of gradient and proximal-gradient methods under the polyak-{\l}ojasiewicz condition.
\newblock {\em Pages  795--811 of:} {\em Joint European Conference on Machine Learning and Knowledge Discovery in Databases}.
\newblock Springer.

\bibitem[\protect\citename{Kempka {\em et~al.}, }2019]{kempka2019adaptive}
Kempka, Michal, Kotlowski, Wojciech, \& Warmuth, Manfred~K. 2019.
\newblock Adaptive scale-invariant online algorithms for learning linear models.
\newblock {\em Pages  3321--3330 of:} {\em International conference on machine learning}.
\newblock PMLR.

\bibitem[\protect\citename{Kingma \& Ba, }2014]{kingma2014adam}
Kingma, Diederik~P, \& Ba, Jimmy. 2014.
\newblock Adam: A method for stochastic optimization.
\newblock {\em arXiv preprint arXiv:1412.6980}.

\bibitem[\protect\citename{Krizhevsky {\em et~al.}, }2009]{krizhevsky2009learning}
Krizhevsky, Alex, Hinton, Geoffrey, {\em et~al.} 2009.
\newblock Learning multiple layers of features from tiny images.

\bibitem[\protect\citename{Li \& Orabona, }2019]{li2019convergence}
Li, Xiaoyu, \& Orabona, Francesco. 2019.
\newblock On the convergence of stochastic gradient descent with adaptive stepsizes.
\newblock {\em Pages  983--992 of:} {\em The 22nd International Conference on Artificial Intelligence and Statistics}.
\newblock PMLR.

\bibitem[\protect\citename{Liu {\em et~al.}, }2023]{liu2023sophia}
Liu, Hong, Li, Zhiyuan, Hall, David, Liang, Percy, \& Ma, Tengyu. 2023.
\newblock Sophia: A Scalable Stochastic Second-order Optimizer for Language Model Pre-training.
\newblock {\em arXiv preprint arXiv:2305.14342}.

\bibitem[\protect\citename{Loshchilov \& Hutter, }2016]{loshchilov2016sgdr}
Loshchilov, Ilya, \& Hutter, Frank. 2016.
\newblock Sgdr: Stochastic gradient descent with warm restarts.
\newblock {\em arXiv preprint arXiv:1608.03983}.

\bibitem[\protect\citename{Loshchilov \& Hutter, }2017]{loshchilov2017decoupled}
Loshchilov, Ilya, \& Hutter, Frank. 2017.
\newblock Decoupled weight decay regularization.
\newblock {\em arXiv preprint arXiv:1711.05101}.

\bibitem[\protect\citename{Luo \& Schapire, }2015]{luo2015achieving}
Luo, Haipeng, \& Schapire, Robert~E. 2015.
\newblock Achieving all with no parameters: Adanormalhedge.
\newblock {\em Pages  1286--1304 of:} {\em Conference on Learning Theory}.
\newblock PMLR.

\bibitem[\protect\citename{Merity {\em et~al.}, }2016]{merity2016pointer}
Merity, Stephen, Xiong, Caiming, Bradbury, James, \& Socher, Richard. 2016.
\newblock Pointer sentinel mixture models.
\newblock {\em arXiv preprint arXiv:1609.07843}.

\bibitem[\protect\citename{Naumov {\em et~al.}, }2019]{naumov2019deep}
Naumov, Maxim, Mudigere, Dheevatsa, Shi, Hao-Jun~Michael, Huang, Jianyu, Sundaraman, Narayanan, Park, Jongsoo, Wang, Xiaodong, Gupta, Udit, Wu, Carole-Jean, Azzolini, Alisson~G, {\em et~al.} 2019.
\newblock Deep learning recommendation model for personalization and recommendation systems.
\newblock {\em arXiv preprint arXiv:1906.00091}.

\bibitem[\protect\citename{Orabona, }2014]{orabona2014simultaneous}
Orabona, Francesco. 2014.
\newblock Simultaneous model selection and optimization through parameter-free stochastic learning.
\newblock {\em Advances in Neural Information Processing Systems}, {\bf 27}.

\bibitem[\protect\citename{Orabona \& P{\'a}l, }2016]{orabona2016coin}
Orabona, Francesco, \& P{\'a}l, D{\'a}vid. 2016.
\newblock Coin betting and parameter-free online learning.
\newblock {\em Advances in Neural Information Processing Systems}, {\bf 29}.

\bibitem[\protect\citename{Orabona \& Tommasi, }2017]{orabona2017training}
Orabona, Francesco, \& Tommasi, Tatiana. 2017.
\newblock Training deep networks without learning rates through coin betting.
\newblock {\em Advances in Neural Information Processing Systems}, {\bf 30}.

\bibitem[\protect\citename{Paquette \& Scheinberg, }2020]{paquette2020stochastic}
Paquette, Courtney, \& Scheinberg, Katya. 2020.
\newblock A stochastic line search method with expected complexity analysis.
\newblock {\em SIAM Journal on Optimization}, {\bf 30}(1), 349--376.

\bibitem[\protect\citename{Paszke {\em et~al.}, }2019]{paszke2019pytorch}
Paszke, Adam, Gross, Sam, Massa, Francisco, Lerer, Adam, Bradbury, James, Chanan, Gregory, Killeen, Trevor, Lin, Zeming, Gimelshein, Natalia, Antiga, Luca, {\em et~al.} 2019.
\newblock Pytorch: An imperative style, high-performance deep learning library.
\newblock {\em Advances in neural information processing systems}, {\bf 32}.

\bibitem[\protect\citename{Rolinek \& Martius, }2018]{rolinek2018l4}
Rolinek, Michal, \& Martius, Georg. 2018.
\newblock L4: Practical loss-based stepsize adaptation for deep learning.
\newblock {\em Advances in neural information processing systems}, {\bf 31}.

\bibitem[\protect\citename{Russakovsky {\em et~al.}, }2015]{russakovsky2015imagenet}
Russakovsky, Olga, Deng, Jia, Su, Hao, Krause, Jonathan, Satheesh, Sanjeev, Ma, Sean, Huang, Zhiheng, Karpathy, Andrej, Khosla, Aditya, Bernstein, Michael, {\em et~al.} 2015.
\newblock Imagenet large scale visual recognition challenge.
\newblock {\em International journal of computer vision}, {\bf 115}, 211--252.

\bibitem[\protect\citename{Schaul {\em et~al.}, }2013]{schaul2013no}
Schaul, Tom, Zhang, Sixin, \& LeCun, Yann. 2013.
\newblock No more pesky learning rates.
\newblock {\em Pages  343--351 of:} {\em International conference on machine learning}.
\newblock PMLR.

\bibitem[\protect\citename{Shazeer \& Stern, }2018]{shazeer2018adafactor}
Shazeer, Noam, \& Stern, Mitchell. 2018.
\newblock Adafactor: Adaptive learning rates with sublinear memory cost.
\newblock {\em Pages  4596--4604 of:} {\em International Conference on Machine Learning}.
\newblock PMLR.

\bibitem[\protect\citename{Tropp {\em et~al.}, }2015]{tropp2015introduction}
Tropp, Joel~A, {\em et~al.} 2015.
\newblock An introduction to matrix concentration inequalities.
\newblock {\em Foundations and Trends{\textregistered} in Machine Learning}, {\bf 8}(1-2), 1--230.

\bibitem[\protect\citename{Vaswani {\em et~al.}, }2019]{vaswani2019painless}
Vaswani, Sharan, Mishkin, Aaron, Laradji, Issam, Schmidt, Mark, Gidel, Gauthier, \& Lacoste-Julien, Simon. 2019.
\newblock Painless stochastic gradient: Interpolation, line-search, and convergence rates.
\newblock {\em Advances in neural information processing systems}, {\bf 32}.

\bibitem[\protect\citename{Ward {\em et~al.}, }2020]{ward2020adagrad}
Ward, Rachel, Wu, Xiaoxia, \& Bottou, Leon. 2020.
\newblock Adagrad stepsizes: Sharp convergence over nonconvex landscapes.
\newblock {\em The Journal of Machine Learning Research}, {\bf 21}(1), 9047--9076.

\bibitem[\protect\citename{Wiseman \& Rush, }2016]{wiseman2016sequence}
Wiseman, Sam, \& Rush, Alexander~M. 2016.
\newblock Sequence-to-sequence learning as beam-search optimization.
\newblock {\em arXiv preprint arXiv:1606.02960}.

\bibitem[\protect\citename{Wu {\em et~al.}, }2018]{wu2018understanding}
Wu, Yuhuai, Ren, Mengye, Liao, Renjie, \& Grosse, Roger. 2018.
\newblock Understanding short-horizon bias in stochastic meta-optimization.
\newblock {\em arXiv preprint arXiv:1803.02021}.

\bibitem[\protect\citename{Zagoruyko \& Komodakis, }2016]{zagoruyko2016wide}
Zagoruyko, Sergey, \& Komodakis, Nikos. 2016.
\newblock Wide residual networks.
\newblock {\em arXiv preprint arXiv:1605.07146}.

\bibitem[\protect\citename{Zhang {\em et~al.}, }2019]{zhang2019lookahead}
Zhang, Michael, Lucas, James, Ba, Jimmy, \& Hinton, Geoffrey~E. 2019.
\newblock Lookahead optimizer: k steps forward, 1 step back.
\newblock {\em Advances in neural information processing systems}, {\bf 32}.

\bibitem[\protect\citename{Zhuang {\em et~al.}, }2020]{zhuang2020adabelief}
Zhuang, Juntang, Tang, Tommy, Ding, Yifan, Tatikonda, Sekhar~C, Dvornek, Nicha, Papademetris, Xenophon, \& Duncan, James. 2020.
\newblock Adabelief optimizer: Adapting stepsizes by the belief in observed gradients.
\newblock {\em Advances in neural information processing systems}, {\bf 33}, 18795--18806.

\end{thebibliography}

\appendix

\section{Proofs from \secref{sec:optimal learning rate}}\label{appen:proofs from optimal LR}

\subsection{Proof of \thmref{thm:optimal LR}}\label{appen:proof of optimal LR}

Using the descent lemma (\thmref{thm:descent lemma}) we get that:
\[
f(\bx_t) - f(\bx_{t+1}) \geq \frac{\inner{\nabla f(\bx_t), \bg_t}^2}{2L\norm{\bg_t}^2}
\]
Taking expectation over the noise, and summing for $t=1,\dots,T$ we get:
\[
\E[f(\bx_1) - f(\bx_T)] \geq \sum_{t=1}^T \E\left[\frac{\inner{\nabla f(\bx_t), \bg_t}^2}{2 L \norm{\bg_t}^2}\right]
\]
Dividing both sides by $T$ and using that $f(\bx^*)$ is a global minimum we get that:
\[
\frac{1}{T}\cdot\E[f(\bx_1) - f(\bx^*)] \geq \frac{1}{T}\cdot\sum_{t=1}^T \E\left[\frac{\inner{\nabla f(\bx_t), \bg_t}^2}{2 L \norm{\bg_t}^2}\right]
\]
Which implies that there is $t_0\in[T]$ such that:
\begin{equation}\label{eq:descent lemma bound}
\E\left[\frac{\inner{\nabla f(\bx_{t_0}), \bg_{t_0}}^2}{2 L \norm{\bg_{t_0}}^2}\right] \leq \frac{2LR}{T}    
\end{equation}

where we used that $\|f(\bx_1)-f(\bx^*)\| \leq R$. The noise is independent at every iteration, hence $\E[\bg_{t_0}|\bx_{t_0}] = \E[\nabla f(\bx_{t_0}) + \bxi_{t_0}|\bx_{t_0}] =  \E[\nabla f(\bx_{t_0})|\bx_{t_0}]$~. Applying this we get:
\begin{align}\label{eq:quad ineq in norm}
    \left(\E[\norm{\nabla f(\bx_{t_0})}^2]\right)^2 &= \left(\E[\inner{\nabla f(\bx_{t_0}), \bg_{t_0}}]\right)^2 \nonumber\\
    & = \left(\E\left[\frac{\inner{\nabla f(\bx_{t_0}), \bg_{t_0}}\norm{\bg_{t_0}}}{\norm{\bg_{t_0}}}\right]\right)^2 \nonumber\\
    & \leq  
    \mathbb{E} \left[\frac{\inner{\nabla f(\bx_{t_0}),\bg_{t_0}}^2}{\norm{\bg_{t_0}}^2}  \right]\cdot \mathbb{E}[\norm{\bg_{t_0}}^2]\nonumber\\
    &\leq
    \frac{2LR}{T} (\mathbb{E}[\|\nabla f(\bx_{t_0})\|^2] + \sigma^2),
\end{align}
where in the first inequality we used Cauchy-Schwartz and in the second inequality we used \eqref{eq:descent lemma bound} and the fact that $\mathbb{E}[\norm{\bg_{t_0}}^2] = \E[\norm{\nabla f(\bx_{t_0}) + \bxi_{t_0}}^2] \leq \mathbb{E}[\|\nabla f(\bx_{t_0})\|^2] + \sigma^2 $ which is true since $\E[\bxi_{t_0}] = 0$ and $\E[\norm{\bxi_{t_0}}^2] \leq \sigma^2$. In total, \eqref{eq:quad ineq in norm} gives us a quadratic inequality in $\E[\norm{\nabla f(\bx_{t_0})}^2]$. Solving this inequality attains:
\[
\E[\norm{\nabla f(\bx_{t_0})}^2] \leq \frac{2LR}{T} + \sigma\sqrt{\frac{2LR}{T}}~.
\]

\subsection{Proof of \thmref{thm:expectation step size}}\label{appen:proof of LR expectation}

We denote $\nabla_t:= \nabla f(\bx_t)$, also denote $\gamma_t:= \frac{\norm{\nabla_t}^2}{\norm{\nabla_t}^2 + \sigma^2}\bg_t$. We denote by $\E_t[\cdot]$ expectation conditioned over $\bxi_1,\dots,\bxi_{t-1}$, for $t=1$ this expectation is unconditioned.
We first have that:
\begin{align}\label{eq:<gamma_t,xi>}
    \E_t[\inner{\gamma_t,\nabla_t}] &= \E_t\left[\inner{\frac{\norm{\nabla_t}^2}{\norm{\nabla_t}^2 + \sigma^2}(\nabla_t + \bxi_t),\nabla_t}\right] \nonumber\\
    & = \frac{\norm{\nabla_t}^4}{\norm{\nabla_t}^2 + \sigma^2} + \E_t\left[\frac{\norm{\nabla_t}^2}{\norm{\nabla_t}^2 + \sigma^2}\inner{\nabla_t,\bxi_t}\right] \nonumber\\
    & = \frac{\norm{\nabla_t}^4}{\norm{\nabla_t}^2 + \sigma^2}
\end{align}
where the last equality is since $\E_t[\bxi_t] = \E[\bxi_t] = 0$. We can also bound:
\begin{align}\label{eq:bound ||gamma_t||}
    \E_t[\norm{\gamma_t}^2] &= \E_t\left[\inner{\frac{\norm{\nabla_t}^2}{\norm{\nabla_t}^2 + \sigma^2}(\nabla_t + \bxi_t), \frac{\norm{\nabla_t}^2}{\norm{\nabla_t}^2 + \sigma^2}\bg_t}(\nabla_t + \bxi_t)\right] \nonumber\\
    & = \left(\frac{\norm{\nabla_t}^2}{\norm{\nabla_t}^2 + \sigma^2}\right)^2\cdot\E_t\left[ \norm{\nabla_t}^2 + 2\inner{\nabla_t,\bxi_t} + \norm{\bxi_t}^2\right] \nonumber\\
    & \leq \left(\frac{\norm{\nabla_t}^2}{\norm{\nabla_t}^2 + \sigma^2}\right)^2 \cdot (\norm{\nabla_t}^2 + \sigma^2) \nonumber\\
    & = \frac{\norm{\nabla_t}^4}{\norm{\nabla_t}^2 + \sigma^2}
\end{align}

Using the descent lemma in expectation over $\bxi_t $  with \eqref{eq:<gamma_t,xi>} and \eqref{eq:bound ||gamma_t||} we get:
\begin{align*}
    \E_t[f(\bx_t) - f(\bx_{t+1})] & \geq \E_t\left[\inner{\nabla_t,\frac{1}{L}\gamma_t} - \frac{L}{2}\left\|\frac{1}{L}\cdot \gamma_t\right\|^2  \right] \\
    & = \frac{1}{L}\E_t[\inner{\nabla_t,\gamma_t}]  - \frac{L}{2L^2}\E_t[\norm{\gamma_t}^2] \\
    & \geq \frac{1}{L}\frac{\norm{\nabla_t}^4}{\norm{\nabla_t}^2 + \sigma^2} - \frac{1}{2L}\frac{\norm{\nabla_t}^4}{\norm{\nabla_t}^2 + \sigma^2} \\
    & =  \frac{1}{2L}\cdot \frac{\norm{\nabla_t}^4}{\norm{\nabla_t}^2 + \sigma^2}
\end{align*}
Summing over all the iterations, dividing by $T$ and taking expectation w.r.t $\bxi_1,\dots,\bxi_T$ we get:
\begin{align*}
    \frac{1}{T}\cdot\E\left[f(\bx_1) - f(\bx_T) \right]  & = \frac{1}{T}\cdot\E\left[\sum_{t=1}^T f(\bx_t) - f(\bx_{t+1})\right] \\
    & = \frac{1}{T}\sum_{t=1}^T\E\left[ f(\bx_t) - f(\bx_{t+1})\right] \\
    & = \frac{1}{T}\sum_{t=1}^T\E\left[\E_t\left[ f(\bx_t) - f(\bx_{t+1})\right]\right] \\
    & \geq \frac{1}{T}\sum_{t=1}^T\E\left[ \frac{1}{2L}\cdot \frac{\norm{\nabla_t}^4}{\norm{\nabla_t}^2 + \sigma^2}\right] ~.
\end{align*}
Note that on the right hand side we have mean over $T$, and the minimum over $t=1,\dots,T$ is smaller than the mean:
\begin{align*}
\min_{t=1,\dots,T}\frac{1}{2L}\cdot\E\left[\frac{\norm{\nabla_t}^4}{\norm{\nabla_t}^2 + \sigma^2}\right] &\leq \frac{1}{T}\cdot\E\left[f(\bx_1) - f(\bx_T) \right] \\
& \leq \frac{1}{T}\cdot\E\left[f(\bx_1) - f(\bx^*) \right]
 \leq \frac{R}{T}
\end{align*}
where $\bx^*$ is a global minimum of $f$, hence $f(\bx_T) \leq f(\bx^*)$. Rearranging the inequality we get:

\begin{equation}\label{eq:y_t bound}
\min_{t=0,\dots,T}\E\left[\frac{\norm{\nabla_t}^4}{\norm{\nabla_t}^2 + \sigma^2}\right] \leq \frac{2LR}{T}    
\end{equation}

Denote the iteration that attains the minimum as $t_0$, and denote $\nabla:= \nabla_{t_0}$. We now have the following:
\begin{align*}
    \left(\E\left[\norm{\nabla}^2\right]\right)^2 &= \left(\E\left[\frac{\norm{\nabla}^2}{\sqrt{\norm{\nabla}^2 + \sigma^2}}\cdot \sqrt{\norm{\nabla}^2 + \sigma^2}\right]\right)^2 \\
    & \leq \E\left[\frac{\norm{\nabla}^4}{\norm{\nabla}^2 + \sigma^2}\right]\cdot \E\left[\norm{\nabla}^2 + \sigma^2\right] \\
    &\leq \frac{2LR}{T}\left(\E\left[\norm{\nabla}^2 \right] + \sigma^2\right)
\end{align*}

where in the first inequality we used Cauchy-Schwartz, and in the second we used \eqref{eq:y_t bound}. In total, we got a quadratic inequality on $\E\left[\norm{\nabla}^2 \right]$. Solving this inequality attains:
\[
\E[\norm{\nabla}^2] \leq \frac{2LR}{T} + \sigma\sqrt{\frac{2LR}{T}}~.
\]


\subsection{Proof of \thmref{thm:sample to approx step size}}\label{appen: proof of sample to approx step size}

We denote $\nabla_t:= \nabla f(\bx_t)$, and also assume w.l.o.g that $M,G, \sigma\geq 1$, otherwise we just replace them in the proofs with $1$. Denote the variance of the noise at iteration $t$ by $\sigma_t:= \E[\norm{\bxi^i_t}^2]$, note that this term is unknown and bounded above by $\sigma^2$. 
To distinguish between the two sets of stochastic gradients the noise vectors as $\bxi_t^i,~\bzeta_t^i$ for $\bh_t^i$ and $\bg_t^i$ respectfully. That is, we write $\bh_t^i = \nabla f(\bx_t) + \bxi_t^i,~ \bg_t^i = \nabla f(\bx_t) + \bzeta_t^i$. 

First note that the estimators $\mu_t$ and $\gamma_t$ can be simplified in the following manner:
\begin{align*}
    &\mu_t = \frac{1}{n(n-1)}\sum_{i\neq j} \inner{\bh^i_t,\bh^j_t} = \frac{1}{n(n-1)}\left(\left\|\sum_{i=1}^n\bh^i_t\right\|^2  - \sum_{i=1}^n \norm{\bh^i_t}^2\right)\\
    & \gamma_t = \frac{1}{n^2}\sum_{i,j=1}^n \inner{\bh^i_t, \bh^j_t} = \frac{1}{n^2}\left\|\sum_{i=1}^n\bh^i_t\right\|^2 
\end{align*}

We first focus on $\gamma_t$, we can write \begin{align}\label{eq:two terms of mu_2}
    \frac{1}{n^2}\left\|\sum_{i=1}^n\bh^i_t\right\|^2  &= \left\| \nabla_t + \frac{1}{n}\sum_{i=1}^n\bxi^i_t\right\|^2\nonumber\\
    & = \norm{\nabla_t}^2 + \frac{2}{n}\sum_{i=1}^n\inner{\nabla_t,\bxi^i_t} + \left\| \frac{1}{n}\sum_{i=1}^n\bxi^i_t\right\|^2
\end{align}

We will show that the two last terms in \eqref{eq:two terms of mu_2} are close to zero w.h.p. For the first term, we rewrite it as $\sum_{i=1}^n\inner{\nabla_t,\frac{2}{n}\bxi^i_t}$. For every $i$, the random variable $\inner{\nabla_t,\frac{2}{n}\bxi^i_t}$ has zero mean, since $\E[\bxi_i^t]=0$. Also by Cauchy-Schwartz, $|\inner{\nabla_t,\frac{2}{n}\bxi^i_t}|\leq \frac{2M G}{n}$. Using Hoeffding's inequality we get:
\begin{align}\label{eq:first event gamma_t}
    P\left(\left|\frac{2}{n}\sum_{i=1}^n\inner{\nabla_t,\bxi^i_t}\right| \geq \frac{\epsilon}{40}\right) &= P\left(\left|\sum_{i=1}^n\inner{\nabla_t,2\bxi^i_t}\right| \geq \frac{n\epsilon}{40}\right) \nonumber\\
    &\leq 2\exp\left(-\frac{2\epsilon^2n^2}{n(160M G)^2}\right) = 2\exp\left(-\frac{2\epsilon^2n}{(160M G)^2}\right) 
\end{align}

For the last term in \eqref{eq:two terms of mu_2}, we will use a matrix version of Bernstein's inequality, see Theorem 6.1.1 in \cite{tropp2015introduction}. We have $\left\|\frac{1}{n}\bxi^i_t\right\|\leq \frac{G}{n}$ and $\E\left[\frac{1}{n}\bxi^i_t\right] = 0$. Denote $Z = \sum_{i=1}^n\frac{1}{n}\bxi^i_t$, to use Bernstein's inequality we need to bound:
\[
\nu(Z):= \max\left\{\left\|\sum_{i=1}^n\frac{1}{n^2}\E[\norm{\bxi^i_t}^2]\right\|, \left\|\sum_{i=1}^n\frac{1}{n^2}\E[\bxi^i_t\bxi_t^{i^\top}]\right\|\right\}
\]
The first term by our assumptions is bounded by $\frac{\sigma^2_t}{n}$. The second term can be bounded in the following way:
\begin{align*}
    \left\|\sum_{i=1}^n\frac{1}{n^2}\E[\bxi^i_t\bxi_t^{i^\top}]\right\| \leq & \frac{1}{n^2}\sum_{i=1}^n\left\|\E[\bxi_t^i\bxi_t^{i^\top}]\right\| \\
    \leq & \frac{1}{n^2}\sum_{i=1}^n\E[\left\|\bxi_t^i\bxi_t^{i^\top}\right\|]
\end{align*}
Where in the last inequality we used Jensen's inequality. Each matrix $\bxi_t^i\bxi_t^{i^\top}$ is rank-1 with an eigenvalue equal to $\E[\norm{\bxi_t^i}^2]$. Hence, in total this term can also be bounded by $\frac{\sigma_t^2}{n}$. Using matrix Bernstein's inequality we get:
\begin{align}\label{eq:second event gamma_t}
    P\left(\left\|\frac{1}{n}\sum_{i=1}^n\bxi_t^i\right\|^2\geq \frac{\epsilon}{40}\right) & = P\left(\left\|\frac{1}{n}\sum_{i=1}^n\bxi_t^i\right\|\geq \sqrt{\frac{\epsilon}{40}}\right) \nonumber\\
    & \leq(d+1) \exp\left(-\frac{\epsilon}{80(\sigma_t^2/n + G\sqrt{\epsilon}/ 3n})\right) \nonumber\\
    & \leq (d+1)\exp\left(-\frac{\epsilon n}{80\sigma_t^2G}\right)
\end{align}

Using a union bound on the events in \eqref{eq:first event gamma_t} and \eqref{eq:second event gamma_t}, we get that there is a universal constant $c_1>0$ such that:
\begin{align*}
    P\left(\left|\gamma_t - \norm{\nabla_t}^2\right|\geq \frac{\epsilon}{20}\right) &= P\left(\left|\frac{2}{n}\sum_{i=1}^n\inner{\nabla_t,\bxi_t^i} + \left\| \frac{1}{n}\sum_{i=1}^n\bxi_t^i\right\|^2\right|\geq \frac{\epsilon}{20}\right) \\
    & \leq 2\exp\left(-\frac{2\epsilon^2n}{(160M G)^2}\right) + (d+1)\exp\left(-\frac{\epsilon n}{80\sigma^2_t G}\right) \\
    & \leq (d+3)\exp\left(-\frac{\epsilon^2 nc_1}{\sigma^2_tM^2 G^2}\right)
\end{align*}

We will now use a similar method to bound $\mu_t$. Note that $\mu_t$ have two terms, both multiplied by a coefficient of $\frac{1}{n(n-1)}$. The second term can be written as:
\begin{align}\label{eq:second term of mu_1}
    \sum_{i=1}^n \norm{\bh_t^i}^2 = \sum_{i=1}^n \norm{\nabla_t + \bxi_t^i}^2 = n\norm{\nabla_t}^2 + \sum_{i=1}^n\inner{\nabla_t, \bxi_t^i} + \sum_{i=1}^n\norm{\bxi_t^i}^2
\end{align}

By our assumption that $n\geq 2$ we have that $\frac{1}{n(n-1)}\left\|\sum_{i=1}^n\bh_t^i\right\|\leq \frac{2}{n^2}\left\|\sum_{i=1}^n\bh_t^i\right\|$. Hence, by a similar analysis as above, there is a constant $c_2>0$ such that:
\begin{align}\label{eq:mu_1 bound first term}
    P\left(\left|\frac{1}{n(n-1)}\left(\left\|\sum_{i=1}^n\bh_t^i\right\|^2 - n\norm{\nabla_t}^2\right) - \norm{\nabla_t}^2\right|\geq \frac{\epsilon}{60}\right) \leq (d+3)\exp\left(-\frac{\epsilon^2 n c_2}{\sigma^2M^2 G^2}\right)
\end{align}

To bound the second of \eqref{eq:second term of mu_1} we use that $\inner{\nabla_t,\bxi_t^i}\leq GM$, $\norm{\bxi_t^i}\leq G$ and Hoeffding's inequality to get:
\begin{align}\label{eq:mu_1 bound second term}
    P\left(\left|\frac{1}{n(n-1)}\sum_{i=1}^n\inner{\nabla_t, \bxi_t^i}\right| \geq \frac{\epsilon}{60}\right) \leq 2\exp\left(-\frac{\epsilon^2n(n-1)^2}{(60GM)^2}\right)
\end{align}

For the third term of \eqref{eq:second term of mu_1} we also use Hoeffding's inequality:
\begin{align}\label{eq:mu_1 bound third term}
    P\left(\left|\frac{1}{n(n-1)}\sum_{i=1}^n\norm{\bxi_t^i}^2 - \frac{\sigma^2_t}{n}\right| \geq \frac{\epsilon}{60}\right) \leq \exp\left(-\frac{\epsilon^2(n-1)^2}{n(60G)^2}\right)
\end{align}
Combining \eqref{eq:mu_1 bound first term}, \eqref{eq:mu_1 bound second term} and \eqref{eq:mu_1 bound third term} we get that there is a constant $c_3>0$ such that:
\begin{align*}
    P\left(\left|\mu_t - \norm{\nabla_t}^2 + \frac{\sigma_t^2}{n}\right|\geq \frac{\epsilon}{20}\right) \leq (d+5)\exp\left(-\frac{\epsilon^2 n c_3}{\sigma_t^2M^2G^2}\right)
\end{align*}
Finally, we combine both probability bounds on $\mu_1^t$ and $\mu_2^t$ to get that there is some constant $c>0$ s.t w.p $> 1- d\exp\left(-\frac{\epsilon^2 n c}{\sigma^2M^2G^2}\right)$ we have both:
\begin{align}
    &\left|\mu_t - \norm{\nabla_t}^2 + \frac{\sigma_t^2}{n}\right|\leq \frac{\epsilon}{20}\label{eq:mu_1 estimation event}\\
    &\left|\gamma_t - \norm{\nabla_t}^2 \right|\leq \frac{\epsilon}{20}\label{eq:mu_2 estimation event}
\end{align}

Note that inside the exponent we replaced $\sigma_t$ with $\sigma$, since this is a lower bound on the probability and $\sigma_t\leq \sigma$. 

We would like to condition on the above events, however these events also depend on the given $\nabla_t$ for iteration $t$, while $\nabla_t$ depends on all the noise vectors from previous iterations. To this end, we denote by $\E_t[\cdot]$ the expectation conditioned on the noise vectors $\bxi_1^i,\dots\bxi_t^i,~\bzeta_1,\dots,\bzeta_{t-1}$ for every $i\in[n]$ and on the events in \eqref{eq:mu_1 estimation event} and \eqref{eq:mu_2 estimation event}. Specifically note that we conditioned on the noise vectors $\bxi_t^i$ including the current iteration, so that we could also condition on the two events above, while on the noise vectors $\bzeta_t^i$ we haven't conditioned on the current iteration, since we would use that it is drawn independently of all the previous noise vectors with zero mean.

We also denote $\alpha_t:= \frac{\mu_t}{\gamma_t}\bg_t$, and we use the assumption that $n\geq \frac{20\sigma^2}{\epsilon}$ to get that:
\begin{align}\label{eq:<gamma_t,xi> approx}
    \E_t[\inner{\alpha_t,\nabla_t}] &= \E_t\left[\inner{\frac{\mu_1^t}{\mu_2^t}\bg_t, \nabla_t}\right]\nonumber\\ 
    &\geq  \E_t\left[\inner{\frac{\norm{\nabla_t}^2  - \frac{\sigma_t^2}{n}- \epsilon/20}{\norm{\nabla_t}^2 + \epsilon/20}\left(\nabla_t + \frac{1}{n}\sum_{i=1}^n\bzeta^i_t\right),\nabla_t}\right] \nonumber\\
    & \geq \E_t\left[\inner{\frac{\norm{\nabla_t}^2 - \epsilon/10}{\norm{\nabla_t}^2 + \frac{\sigma_t^2}{n} + \epsilon/20}\left(\nabla_t + \frac{1}{n}\sum_{i=1}^n\bzeta^i_t\right),\nabla_t}\right] \nonumber\\
    &\geq \frac{18}{21}\E_t\left[\inner{\frac{\norm{\nabla_t}^2}{\norm{\nabla_t}^2  + \frac{{\sigma}_t^2}{n}}\left(\nabla_t + \frac{1}{n}\sum_{i=1}^n\bzeta^i_t\right),\nabla_t}\right] \nonumber\\
    & = \frac{18}{21}\cdot\frac{\norm{\nabla_t}^4}{\norm{\nabla_t}^2 + \frac{{\sigma}_t^2}{n}} +\frac{18}{21n}\sum_{i=1}^n \E_t\left[\frac{\norm{\nabla_t}^2}{\norm{\nabla_t}^2 + \frac{{\sigma}_t^2}{n}}\inner{\nabla_t,\bzeta^i_t}\right] \nonumber\\
    & = \frac{18}{21}\cdot\frac{\norm{\nabla_t}^4}{\norm{\nabla_t}^2 + \frac{{\sigma}_t^2}{n}}
\end{align}

where we used that $\E_t[\bzeta_t^i] = \E[\bzeta_t^i] = 0$. We also have that:
\begin{align}\label{eq:bound ||gamma_t|| approx}
    \E_t[\norm{\alpha_t}^2] &= \E_t\left[\inner{\frac{\mu_t}{\gamma_t}\left(\nabla_t + \frac{1}{n}\sum_{i=1}^n\bzeta^i_t\right), \frac{\mu_t}{\gamma_t}\left(\nabla_t + \frac{1}{n}\sum_{i=1}^n\bzeta^i_t\right)}\right] \nonumber\\
    & = \left(\frac{\mu_t}{\gamma_t}\right)^2\cdot\E_t\left[ \norm{\nabla_t}^2 + \frac{2}{n}\sum_{i=1}^n\inner{\nabla_t,\bzeta^i_t} + \left\|\frac{1}{n}\sum_{i=1}^n\bzeta^i_t\right\|^2\right] \nonumber\\
    & \leq \left(\frac{\norm{\nabla_t}^2 - \frac{\sigma_t^2}{n} + \epsilon/20}{\norm{\nabla_t}^2 - \epsilon/20}\right)^2 \cdot \left(\norm{\nabla_t}^2 + \frac{{\sigma}_t^2}{n}\right) \nonumber\\
    & \leq \left(\frac{\norm{\nabla_t}^2 + \epsilon/20}{\norm{\nabla_t}^2 + \frac{\sigma^2}{n}- \epsilon/10}\right)^2 \cdot \left(\norm{\nabla_t}^2 + \frac{{\sigma}_t^2}{n}\right) \nonumber\\
    & \leq \left(\frac{21}{18}\right)^2\cdot \left(\frac{\norm{\nabla_t}^2}{\norm{\nabla_t}^2 + \frac{{\sigma}_t^2}{n}}\right)\cdot \left(\norm{\nabla_t}^2 + \frac{{\sigma}_t^2}{n}\right)\nonumber\\
    & = \left(\frac{21}{18}\right)^2\frac{\norm{\nabla_t}^4}{\norm{\nabla_t}^2 + \frac{{\sigma}_t^2}{n}}
\end{align}

Combining \eqref{eq:<gamma_t,xi> approx} and \eqref{eq:bound ||gamma_t|| approx} and the descent lemma we get that:

\begin{align*}
    \E_t[f(\bx_t) - f(\bx_{t+1})] & \geq \E_t\left[\inner{\nabla_t,\frac{1}{L}\alpha_t} - \frac{L}{2}\left\|\frac{1}{L}\cdot \alpha_t\right\|^2  \right] \\
    & = \frac{1}{L}\E_t[\inner{\nabla_t,\alpha_t}]  - \frac{L}{2L^2}\E_t[\norm{\alpha_t}^2] \\
    & \geq \frac{18}{21 L}\frac{\norm{\nabla_t}^4}{\norm{\nabla_t}^2 + \frac{{\sigma}_t^2}{n}} - \frac{441}{648L}\frac{\norm{\nabla_t}^4}{\norm{\nabla_t}^2 + \frac{{\sigma}_t^2}{n}} \\
    & \geq   \frac{1}{6L}\cdot \frac{\norm{\nabla_t}^4}{\norm{\nabla_t}^2 + \frac{{\sigma}_t^2}{n}}\\
    & \geq   \frac{1}{6L}\cdot \frac{\norm{\nabla_t}^4}{\norm{\nabla_t}^2 + \frac{\sigma^2}{n}}
\end{align*}
where in the last inequality we used that $\sigma_t \leq \sigma$. We now follow in the same manner as in \thmref{thm:optimal LR} where we replace $\sigma^2$ by $\frac{\sigma^2}{n}$. Namely, we sum over all the iteration $t\in[T]$, divide by $T$ and take expectation w.r.t all the noise vectors $\bxi_t^i,~\bzeta_t^i$ for $t\in[T],~i\in[n]$ to get:
\begin{align*}
    \frac{1}{T}\cdot\E\left[f(\bx_1) - f(\bx_T) \right]  & = \frac{1}{T}\cdot\E\left[\sum_{t=1}^T f(\bx_t) - f(\bx_{t+1})\right] \\
    & = \frac{1}{T}\sum_{t=1}^T\E\left[ f(\bx_t) - f(\bx_{t+1})\right] \\
    & = \frac{1}{T}\sum_{t=1}^T\E\left[\E_t\left[ f(\bx_t) - f(\bx_{t+1})\right]\right] \\
    & \geq \frac{1}{T}\sum_{t=1}^T\E\left[ \frac{1}{6L}\cdot \frac{\norm{\nabla_t}^4}{\norm{\nabla_t}^2 + \frac{\sigma^2}{n}}\right]~.
\end{align*}

On the right hand side of the above inequality we have mean over $T$, and the minimum over $t=1,\dots,T$ is smaller than the mean:
\begin{align*}
\min_{t=1,\dots,T}\frac{1}{6L}\cdot\E\left[\frac{\norm{\nabla_t}^4}{\norm{\nabla_t}^2 + \frac{\sigma^2}{n}}\right] &\leq \frac{1}{T}\cdot\E\left[f(\bx_1) - f(\bx_T) \right] \\
& \leq \frac{1}{T}\cdot\E\left[f(\bx_1) - f(\bx^*) \right]
 \leq \frac{R}{T}
\end{align*}
where $\bx^*$ is a global minimum of $f$, hence $f(\bx_T) \leq f(\bx^*)$. Rearranging the inequality we get:

\begin{equation}\label{eq:y_t bound2}
\min_{t=1,\dots,T}\E\left[\frac{\norm{\nabla_t}^4}{\norm{\nabla_t}^2 + \frac{\sigma^2}{n}}\right] \leq \frac{6LR}{T}~.
\end{equation}

Denote the iteration that attains the minimum as $t_0$, and denote $\nabla:= \nabla_{t_0}$. We now have the following:
\begin{align*}
    \left(\E\left[\norm{\nabla}^2\right]\right)^2 &= \left(\E\left[\frac{\norm{\nabla}^2}{\sqrt{\norm{\nabla}^2 + \frac{\sigma^2}{n}}}\cdot \sqrt{\norm{\nabla}^2 + \frac{\sigma^2}{n}}\right]\right)^2 \\
    & \leq \E\left[\frac{\norm{\nabla}^4}{\norm{\nabla}^2 + \frac{\sigma^2}{n}}\right]\cdot \E\left[\norm{\nabla}^2 + \frac{\sigma^2}{n}\right] \\
    &\leq \frac{6LR}{T}\left(\E\left[\norm{\nabla}^2 \right] + \frac{\sigma^2}{n}\right)
\end{align*}

where in the first inequality we used Cauchy-Schwartz, and in the second we used \eqref{eq:y_t bound2}. In total, we got a quadratic inequality on $\E\left[\norm{\nabla}^2 \right]$. Solving this inequality attains:

\[
\E[\norm{\nabla}^2] \leq \frac{6LR}{T} + \frac{\sigma}{\sqrt{n}}\sqrt{\frac{6LR}{T}}~.
\]

In particular, by our choice of $n= \Omega\left(\frac{\sigma^2}{\epsilon}\right)$ we get that after $T=O\left(\frac{1}{\epsilon}\right)$ iterations we achieve $\E[\norm{\nabla}^2] \leq \epsilon$. Recall that this convergence result is conditioned on the events in \eqref{eq:mu_1 estimation event} and \eqref{eq:mu_2 estimation event}, which happens w.p.\ $> 1- dT\exp\left(-\frac{\epsilon^2 n c}{\sigma^2M^2G^2}\right)$. In total, we get the convergence result w.h.p.\ by choosing $n=\Omega\left(\frac{\sigma^2M^2G^2\log(d)}{\epsilon^2}\right)$, where we achieve $\E[\norm{\nabla}^2] \leq \epsilon$ by receiving a total of $O\left(\frac{\sigma^2M^2 G^2\log(d)}{\epsilon^3}\right)$ noisy gradients.

\section{Proof from \secref{sec:practical implementation}}\label{appen:proofs from practical}

\subsection{Proof of \thmref{thm:variance of the norm}}

We first have that:
\begin{align*}
    \E[\mu] & = \frac{1}{n(n-1)}\sum_{i\neq j}\E[\inner{\bg_i,\bg_j}]\\
    & = \frac{1}{n(n-1)}\sum_{i\neq j}\E[\inner{\nabla + \bxi_i,\nabla + \bxi_j}] \\
    & = \norm{\nabla} + \frac{1}{n(n-1)}\sum_{i\neq j}\E[\inner{\bxi_i, \bxi_j}]  = \norm{\nabla}
\end{align*}
where we used that the $\bxi_i$'s are independent with zero mean. For the variance calculation, we have:
\begin{align*}
    \var(\mu) & = \var\left(\frac{1}{n(n-1)}\sum_{i\neq j}\inner{\bg_{i}, \bg_{j}}\right) \\
    & = \frac{1}{n^2(n-1)^2}\var\left(\sum_{i\neq j} \inner{\nabla + \bxi_i, \nabla + \bxi_j}\right)\\
    & = \frac{1}{n^2(n-1)^2}\var\left(n(n-1)\norm{\nabla} + \sum_{i\neq j} \inner{\nabla,\bxi_i} + \inner{\nabla,\bxi_j} +   \inner{ \bxi_i, \bxi_j}\right) \\
    & = 
    \frac{1}{n^2(n-1)^2}\var\left(2(n-1)\sum_{i=1}^n\inner{\nabla,\bxi_i} + \sum_{i\neq j}\inner{ \bxi_i, \bxi_j} \right)
\end{align*}
Here, we used that $\nabla$ is a constant (non-random) vector. We calculate this variance by extending it as an expectation:
\begin{align*}
    \var(\mu) &= \frac{1}{n^2(n-1)^2}\Bigg(\E\left[\left(2(n-1)\sum_{i=1}^n\inner{\nabla,\bxi_i} + \sum_{i\neq j}\inner{ \bxi_i, \bxi_j} \right)^2\right] \\
    & - \left(\E\left[2(n-1)\sum_{i=1}^n\inner{\nabla,\bxi_i} + \sum_{i\neq j}\inner{ \bxi_i, \bxi_j} \right]\right)^2\Bigg)
\end{align*}
We will calculate each term of the above separately. For the second term, note that $\E[\inner{\nabla,\bxi_i}] =0$, and for $i\neq j: ~$  $\E[\inner{\bxi_i,\bxi_j}] =0$. Hence, by the linearity of the expectation, the second term is equal to zero. The first term (without the coefficient) is equal to:
\begin{align}
    &\E\left[\left(2(n-1)\sum_{i=1}^n\inner{\nabla,\bxi_i} + \sum_{i\neq j}\inner{ \bxi_i, \bxi_j} \right)^2\right] \nonumber\\
    =& \E\left[4(n-1)^2\left(\sum_{i=1}^n\inner{\nabla,\bxi_i}\right)^2\right] + \E\left[\left(\sum_{i\neq j}\inner{\bxi_i,\bxi_j}\right)^2\right] + \nonumber\\
    & + \E\left[4(n-1)\left(\sum_{i=1}^n\inner{\nabla,\bxi_i}\right) \cdot \left(\sum_{k\neq j}\inner{\bxi_k,\bxi_j}\right)\right]\label{eq:three terms variance}
\end{align}

For the first term in \eqref{eq:three terms variance} we have:
\begin{align}
    \E\left[4(n-1)^2\left(\sum_{i=1}^n\inner{\nabla,\bxi_i}\right)^2\right] &= 4(n-1)^2\E\left[\sum_{i=1}^n\sum_{j=1}^n\inner{\nabla,\bxi_i}\cdot\inner{\nabla,\bxi_j}\right] \nonumber\\
    & = 4(n-1)^2\E\left[\sum_{i=1}^n\inner{\nabla,\bxi_i}^2\right]\nonumber\\
    &\leq 4(n-1)^2\sum_{i=1}^n\E\left[\norm{\nabla^2}\cdot\norm{\bxi_i}^2\right]\label{eq:first inequality variance bound}\\
    & = 4n(n-1)^2\norm{\nabla}^2\sigma^2 \nonumber
\end{align}
where we used Cauchy-Schwartz and that the $\bxi_i$'s are independent with zero mean, hence $\E[\inner{\nabla,\bxi_i}\cdot\inner{\nabla,\bxi_j}] = 0$ for $i\neq j$. For the second term in \eqref{eq:three terms variance} we have:

\begin{align}
    \E\left[\left(\sum_{i\neq j}\inner{\bxi_i,\bxi_j}\right)^2\right] &= \E\left[\left(\sum_{i\neq j}\inner{\bxi_i,\bxi_j}\right)\cdot \left(\sum_{k\neq M}\inner{\bxi_k,\bxi_M}\right)\right] \nonumber\\
    & = \sum_{i\neq j}\sum_{k\neq M}\E\left[\inner{\bxi_i,\bxi_j}\cdot \inner{\bxi_k,\bxi_M}\right] \nonumber\\
    & = \sum_{i\neq j}\E\left[\inner{\bxi_i,\bxi_j}^2\right] \nonumber\\
    & \leq  \sum_{i\neq j}\E[\norm{\bxi_i}^2\cdot \norm{\bxi_j}^2] \label{eq:second inequality variance bound}\\
    &= {n(n-1)}\sigma^4\nonumber
\end{align}
For the third term in \eqref{eq:three terms variance} we get:

\begin{align*}
    &E\left[4(n-1)\left(\sum_{i=1}^n\inner{\nabla,\bxi_i}\right) \cdot \left(\sum_{k\neq j}\inner{\bxi_k,\bxi_j}\right)\right] \\
    = & 4(n-1)\sum_{i=1}^n\sum_{k\neq j}\E\left[\inner{\nabla,\bxi_i} \cdot \inner{\bxi_k,\bxi_j}\right] = 0
\end{align*}
where again we used the the $\bxi_i$'s are independent with zero mean. Combining the three terms above, we get that:
\begin{align*}
    \var(\mu)\leq \frac{4}{n}\norm{\nabla}^2\sigma^2 + \frac{1}{n(n-1)} \sigma^4
\end{align*}
In particular, if we assume that $\bxi_i\sim \Ncal\left(0, \frac{\sigma^2}{d}I_d\right)$, then instead of the inequalities in \eqref{eq:first inequality variance bound} and \eqref{eq:second inequality variance bound} we can derive equalities. For \eqref{eq:first inequality variance bound} we have:

\begin{align*}
    4(n-1)^2\E\left[\sum_{i=1}^n\inner{\nabla,\bxi_i}^2\right] = 4n(n-1)^2\norm{\nabla}^2\frac{\sigma^2}{d}
\end{align*}
where we used that $\bxi_i$ have a spherically symmetric distribution, hence $\inner{\nabla,\bxi_i}$ have the same distribution for any fixed $\nabla$, then can assume w.l.o.g that $\nabla = \norm{\nabla}\cdot \be_1$. For \eqref{eq:second inequality variance bound} we have 

\begin{align*}
    \sum_{i\neq j}\E\left[\inner{\bxi_i,\bxi_j}^2\right] &= \sum_{i\neq j}\E\left[\left(\sum_{k=1}^d\xi_{i,k}\xi_{j,k}\right)^2\right]\\
    & = \sum_{i\neq j}\E\left[\sum_{k=1}^d\xi_{i,k}^2\xi_{j,k}^2\right] \\
    & = \sum_{i\neq j}\sum_{k=1}^d\E[\xi_{i,k}^2]\E[\xi_{j,k}^2]\\
    & = \frac{\sigma^4}{dn(n-1)}
\end{align*}
where we used that each coordinate is distributed i.i.d. Summing everything together, we have that:
\begin{align*}
    \var(\mu) = \frac{4\norm{\nabla}^2\sigma^2}{nd} + \frac{\sigma^4}{dn(n-1)}
\end{align*}

\section{Additional Details for the Practical Implementation of GLyDER}\label{appen:additional practical details}

\paragraph{Efficient implementation of the inner products.} Note that to efficiently estimate the norm using \eqref{eq: unbiased norm estimator} with $n$ stochastic gradients we need to do $O(n^2)$ inner products. Instead, we can use the formulas:
\[
\left\|\sum_{i=1}^n\bg^i_t\right\|^2 - \sum_{i=1}^n\left\|\bg^i_t\right\|^2 = \sum_{i\neq j} \inner{\bg^i_t,\bg^j_t}~,~~~\left\|\sum_{i=1}^n\bg^i_t\right\|^2 = \sum_{i,j = 1}^n \inner{\bg^i_t,\bg^j_t}
\]
which requires performing only $O(n)$ operations. We note that these estimators could also be used in \algref{alg:greedy constant L} and the analysis would remain the same, however we decided to present it as inner products to make the presentation clearer.

Also note that the estimator of those terms (as done in \thmref{thm:sample to approx step size}) may be negative, although they should be non-negative as they estimate the norm of the gradient and the variance of the noise. To overcome this deviation, we clip the learning rate at $0$, i.e. if the term is negative we define the learning rate at this iteration to be $0$.

\paragraph{Utilizing parallel computational units.} Common modern machine learning libraries (e.g. PyTorch \cite{paszke2019pytorch}, Tensorflow \cite{abadi2016tensorflow}) calculate the gradient for an entire batch of samples, instead of separately for each sample. This means, that in practice even if we are given a batch of samples, we receive only a single stochastic gradient which is an aggregation (sum or mean) of the gradient of the loss on each sample. 

A common practice in modern machine learning application is using parallelized computation, where each computational node receive only a part of the batch. This is done e.g. when using TPUs \citep{jouppi2017datacenter}, which is a commonly used processing unit for training neural networks. We can model it as if we are given samples $\bx_1,\dots,\bx_m$, and we have $k \ll m$ processing units, each given $\frac{m}{k}$ samples (assume for simplicity that $\frac{m}{k}$ is an integer). The processing units calculate in parallel the stochastic gradients $\bh_1,\dots,\bh_k$, each w.r.t its own batch of samples, and aggregates them. To estimate the learning rate, we can use the $\bh_i$'s in \eqref{eq: unbiased norm estimator}. Note that according to \thmref{thm:variance of the norm}, using these gradients will have the same bound on the variance (up to a constant) than calculating the gradient w.r.t each sample separately and estimating the norm using $m$ stochastic gradients instead of $k$.

\paragraph{Initial learning rate.}
Our learning rate scheduler in theory does not require an initial learning rate. However, adding exponential averaging to reduce the noise does require an initial learning rate, otherwise the first iterations (which may be very noisy) will have a very significant effect on performance of the algorithm. We add a tunable parameter $\eta_0$ which will be the initial learning rate. We emphasize that this is the only parameter in our learning rate scheduler which is being tuned.

\subsection{Extension of GLyDER to Other Optimizers}\label{appen:greedy extension}

We provide an extension of the GLyDER stepsize scheduler to general optimization algorithms in \algref{alg:greedy other optimizers}, where we focus only on the first option for smoothness estimation in Subsection \ref{subsec:smoothness estimation}, i.e. projection to a $1$-dimensional function. Our assumption of the algorithm is that at each iteration $t$ the algorithm outputs a descent direction $\bd_t$. This descent direction might not be the gradient of the objective, although it often depend on the gradient in some manner. Typical examples include momentum SGD, where the descent direction is the gradient plus some momentum term which depends on the gradients from previous iterations. Other examples include per-parameter algorithms such as Adam or Adagrad, where the descent direction is calculated using the gradient of each parameter separately.

We additionally assume that we have access to the stochastic second derivative of the objective $f(\cdot)$ at $\bx_{t-1}$ projected on a given direction. We now explain why this is a reasonable assumption which is applicable to practically all use-cases in supervised learning: At each iteration in supervised learning under a stochastic setting the algorithm receives a batch of labeled samples, and calculates the gradient based on the loss on those samples. In other words, the objective function is the loss on those batch of samples, and the stochastic gradient is the gradient of this loss function. Thus, calculating the second derivative of the objective $f(\cdot)$ is done in a similar way, and w.r.t the batch of samples. We emphasize that calculating the second derivative \emph{projected} on a given direction can be done in time complexity similar to that of calculating the gradient itself.

\begin{algorithm}[H]\label{alg:greedy other optimizers}
 \textbf{Input:} $\bx_0,~ n, ~\eta_0, \beta$
 
 \For{$t=1,2,\dots,T$}{
  \textbf{Sample} stochastic gradients $\bg_t^1,\dots,\bg_t^n$\\
  \textbf{Set:} \\
  ~~~~ $\mu_t:= \left\|\sum_{i=1}^n\bg^i_t\right\|^2 - \sum_{i=1}^n\left\|\bg^i_t\right\|^2$ \\
  ~~~~ $\gamma_t:= \left\|\sum_{i=1}^n\bg^i_t\right\|^2 $ \\
  ~~~~ $\bg_t = \sum_{i=1}^n\bg_t^i$
  
  \If {$\gamma_t = 0$ or $\mu_t \leq 0$}{Set $\frac{\mu_t}{\gamma_t}:= 0$}
  
  Receive a descent direction $\bd_t$ \\
  estimate $L_t$ using a projection to a $1$-dimensional function, projected on the direction $\bd_t$
  
  \textbf{Set:} $\eta_t := (1-\beta)\eta_{t-1} + \beta\cdot\frac{1}{L_t}\cdot \frac{\mu_t}{\gamma_t}$
  
  \textbf{Update} $\bx_t = \bx_{t-1} - \eta_t\bd_t$
  }
 \caption{GLyDER scheduler for general optimization algorithms}
\end{algorithm}

\subsection{GLyDeR Stepsize Examples}

In \figref{fig:lr_hessian_gnb} we show the learning rate from \algref{alg:greedy vary L} when training on the CIFAR100 dataset, for different initial learning rates. We observe a "warmup-like" behavior in the first iterations, and then a decrease where the speed and intensity of the decrease depends on the initial learning rate. This behavior might be due to the exponential averaging.

\begin{figure}[ht]
    \centering
    \includegraphics[width=0.45\linewidth]{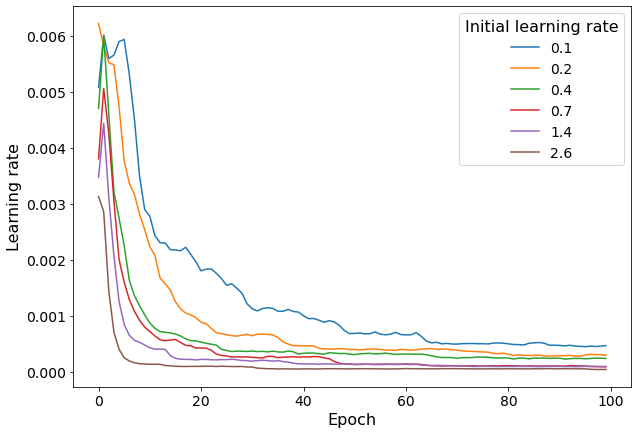}
    \includegraphics[width=0.45\linewidth]{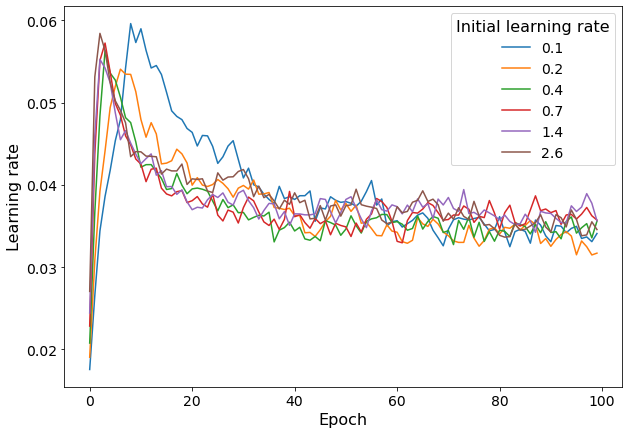}
    \caption{The GLyDeR learning rate scheduler trained on CIFAR100 for different initial learning rates. \textbf{Left:} Smoothness estimation using projection to a $1$-dimensional function (option $1$). \textbf{Right:} Smoothness estimation using GNB estimation (option $2$).}
    \label{fig:lr_hessian_gnb}
\end{figure}

\section{Experimental Details}\label{appen: exp details}

In the following section we will detail the full experimental details for all the experiments performed in the paper. We emphasize that beyond the stepsize scheduler and choice of optimization algorithm, we used the default hyper-parameter choice from the init2winit framework on all the experiments. All the parameters are chosen as to optimize the performance on each task separately.

\subsection*{Datasets}

\paragraph{CIFAR10/100.} We used a wide-RedNet architecture, batch size of $128$, and trained for $300$ epochs.

\paragraph{Imagenet.} We used a ResNet50 architecture with batch size of $512$, and trained for $100$ epochs.

\paragraph{WikiText-2} We used an LSTM model, with an embedding dimension of $200$, batch size of $32$ and trained for $500000$ steps.

\paragraph{Criteo} We used a DLRM model with and bottom MLP dimensions of $(512, 256, 128)$ and top MLP dimensions of $(1024, 1024, 512, 256, 1)$. The embedding dimension is $128$, batch size of $524,288$ and trained for $1.3$ epochs.

\subsection*{Optimizers}
For momentum SGD we used a momentum parameter of $0.9$. For Adam we used $\beta_1 = 0.9,~\beta_2=0.999$ and $\epsilon = 10^{-7}$. For vanilla SGD there we no hyper-parameters beyond these of the stepsize scheduler. For all the experiments and all the schedulers we used the exact same parameters.

\subsection*{Hyper-parameter search} 
\paragraph{Initial stepsize:} We employ a set of $20$ different stepsizes, evenly distributed in a logarithmic scale spanning from $10^{-3}$ to $10^2$ in base $10$. These $20$ stepsizes were used for all the schedulers and algorithms.

\paragraph{Squash steps for rsqrt:} We employed the following squash steps for the rsqrt scheduler: $0.5,~ 1, 5, 10, 15, 50, 100, 200, 500, 1000$.

\paragraph{Number of epochs:} We refrained from conducting a hyper-parameter search for the number of epochs, since altering the training duration can substantially impact the impartiality of the outcome. However, we note that we used the default number of epoch from the init2winit framework, which is optimized for the default scheduler used in each specific dataset, in our cases either cosine or rsqrt. Hence, in practice for each dataset we used the number of epochs which is optimized to work on one of the schedulers which is compared to GLyDER.

\section{Additional Experiments}\label{appen: additional experiments} In addition to the experiments shown in Table \ref{tab:results momentum}, we have performed similar experiments when training with Adam and vanilla SGD. The experiments are shown in Table \ref{tab:results sgd} and Table \ref{tab:results adam}. The GLyDER scheduler is comparative with the other manually tuned schedulers up to a small error. Note that there is no one scheduler which outperforms all the others, e.g. cosine decay performs well on most tasks but also under-performs on a few of them (e.g. Criteo with SGD and CIFAR100 with Adam). Thus, choosing the right scheduler for the task can be seen as an extra hyper-parameter that requires tuning.

In Table \ref{tab:perplexity} we provide the perplexity of the experiments done with the WikiText-2 dataset, over all the schedulers and optimization algorithms.

\begin{table}[ht]
\begin{tabular}{|l|lllll|}
\hline
                              & \multicolumn{1}{l|}{\textbf{GLyDER + $1$-- d proj}} & \multicolumn{1}{l|}{\textbf{GLyDER + GNB}} & \multicolumn{1}{l|}{\textbf{Constant}} & \multicolumn{1}{l|}{\textbf{Cosine}}  & \textbf{rsqrt}                                                                                                                                                                                  \\ \hline
\textbf{CIFAR10 $\downarrow$}              & \multicolumn{1}{l|}{$4.2 \% \pm 0.03$}            & \multicolumn{1}{l|}{$4.4 \%\pm 0.2$}         & \multicolumn{1}{l|}{$4.5\% \pm 0.3$}     & \multicolumn{1}{l|}{$3.0\% \pm 0.2$}    & $4.4 \%\pm 0.1$                \\ \hline
\textbf{CIFAR100 $\downarrow$}             & \multicolumn{1}{l|}{$19.6\% \pm 0.2$}            & \multicolumn{1}{l|}{$22.6 \%\pm 0.6$}        & \multicolumn{1}{l|}{$23.9 \%\pm 0.6$}    & \multicolumn{1}{l|}{$18.8 \%\pm 0.1$}   & $21.5 \% \pm 0.2$              \\ \hline
\textbf{Imagenet $\downarrow$}             & \multicolumn{1}{l|}{$24.8\% \pm 0.5$}            & \multicolumn{1}{l|}{$34.9 \%\pm 0.5$}        & \multicolumn{1}{l|}{$32.0\% \pm 0.2$}    & \multicolumn{1}{l|}{$23.6 \%\pm 0.03$}  & $28.8\% \pm 0.3$               \\ \hline
\textbf{WikiText-2 $\downarrow$}        & \multicolumn{1}{l|}{$78.9 \%\pm 0.9$}            & \multicolumn{1}{l|}{$79.9\% \pm 0.1$}      & \multicolumn{1}{l|}{$75.9 \%\pm 0.0$}    & \multicolumn{1}{l|}{$76.0\% \pm 0.02$}  & $75.9 \%\pm 0.1$               \\ \hline
\textbf{Criteo $\uparrow$}               & \multicolumn{1}{l|}{$0.76 \pm 0.003$}          & \multicolumn{1}{l|}{$0.68 \pm 0.005$}      & \multicolumn{1}{l|}{$0.74 \pm 0.001$}  & \multicolumn{1}{l|}{$0.72 \pm 0.003$} & $0.7 \pm 0.003$              \\ \hline

\end{tabular}
\caption{Similar to Table \ref{tab:results momentum}, except that here the experiments are done using vanilla SGD.}
\label{tab:results sgd}
\end{table}

\begin{table}[ht]
\begin{tabular}{|l|lllll|}
\hline
                              & \multicolumn{1}{l|}{\textbf{GLyDER + $1$-- d proj}} & \multicolumn{1}{l|}{\textbf{GLyDER + GNB}} & \multicolumn{1}{l|}{\textbf{Constant}} & \multicolumn{1}{l|}{\textbf{Cosine}}  & \textbf{rsqrt}               \\ \hline

\textbf{CIFAR10 $\downarrow$}              & \multicolumn{1}{l|}{$3.6 \%\pm 0.5$}             & \multicolumn{1}{l|}{$4.3\%\pm 0.1$}         & \multicolumn{1}{l|}{$8.2\% \pm 0.5$}     & \multicolumn{1}{l|}{$5.3\% \pm 0.3$}    & $4.6 \%\pm 0.2$                \\ \hline
\textbf{CIFAR100 $\downarrow$}             & \multicolumn{1}{l|}{$19.3 \%\pm 0.4$}            & \multicolumn{1}{l|}{$20.5\% \pm 0.3$}        & \multicolumn{1}{l|}{$27.7 \%\pm 0.3$}    & \multicolumn{1}{l|}{$21.5 \%\pm 0.1$}   & $21.7 \%\pm 0.5$               \\ \hline
\textbf{Imagenet $\downarrow$}             & \multicolumn{1}{l|}{$41.5 \%\pm 0.5$}            & \multicolumn{1}{l|}{$29.6\% \pm 0.07$}       & \multicolumn{1}{l|}{$43.7 \%\pm 0.7$}    & \multicolumn{1}{l|}{$29.1 \%\pm 0.09$}  & $28.3 \%\pm 0.2$               \\ \hline
\textbf{WikiText-2 $\downarrow$}        & \multicolumn{1}{l|}{$96.1 \%\pm 2.0$}            & \multicolumn{1}{l|}{$77.8 \%\pm 0.4$}        & \multicolumn{1}{l|}{$77.0\% \pm 0.02$}   & \multicolumn{1}{l|}{$77.3 \%\pm 1.1$}   & $76.6 \%\pm 0.09$              \\ \hline
\textbf{Criteo $\uparrow$}               & \multicolumn{1}{l|}{$0.78 \pm 0.0$}            & \multicolumn{1}{l|}{$0.78 \pm 0.03$}       & \multicolumn{1}{l|}{$0.78 \pm 0.0$}    & \multicolumn{1}{l|}{$0.78 \pm 0.0$}   & $0.79 \pm 0.004$             \\ \hline
\end{tabular}
\caption{Similar to Table \ref{tab:results momentum}, except that here the experiments are done using the  Adam optimizer.}
\label{tab:results adam}
\end{table}

\begin{table}[ht]
\begin{tabular}{|l|lllll|}
\hline
                              & \multicolumn{1}{l|}{\textbf{GLyDER + $1$-- d proj}} & \multicolumn{1}{l|}{\textbf{GLyDER + GNB}} & \multicolumn{1}{l|}{\textbf{Constant}} & \multicolumn{1}{l|}{\textbf{Cosine}}  & \textbf{rsqrt}                                                                                                                                                                                  \\ \hline
\textbf{SGD}              & \multicolumn{1}{l|}{$245 \pm 32$}            & \multicolumn{1}{l|}{$277 \pm 2$}         & \multicolumn{1}{l|}{$151 \pm 0.6$}     & \multicolumn{1}{l|}{$153.\pm 0.9$}    & \multicolumn{1}{l|}{$152 \pm 2.4$}                \\ \hline
\textbf{Momentum SGD}             & \multicolumn{1}{l|}{$177 \pm 2$}            & \multicolumn{1}{l|}{$172 \pm 1$}        & \multicolumn{1}{l|}{$148 \pm 0.4$}    & \multicolumn{1}{l|}{$149 \pm 0.4$}   & \multicolumn{1}{l|}{$153
2 \pm 0.2$}              \\ \hline
\textbf{Adam}             & \multicolumn{1}{l|}{$5659 \pm 509$}            & \multicolumn{1}{l|}{$256 \pm 21$}        & \multicolumn{1}{l|}{$2002 \pm 48$}    & \multicolumn{1}{l|}{$3567 \pm 72$}  & \multicolumn{1}{l|}{$211 \pm 2.1$}               
\\\hline
\end{tabular}
\caption{Perplexity for the experiments with the WikiText-2 dataset.}
\label{tab:perplexity}
\end{table}

\end{document}